\documentclass[10pt,journal,cspaper,compsoc, final]{IEEEtran}
\usepackage[T1]{fontenc}
\usepackage{algorithm}
\usepackage{algorithmic}
\usepackage{adjustbox}
\usepackage{multirow}

\usepackage{comment}
\usepackage{hyperref}
\usepackage{url}
\usepackage{xcolor,colortbl}
\usepackage{amsmath}
\usepackage{bm}
\usepackage{amsfonts}
\usepackage{amssymb}
\usepackage{subfigure}
\usepackage{mathtools}
\usepackage[english]{babel}
\usepackage[inline]{enumitem}
\usepackage{setspace}

\ifCLASSOPTIONcompsoc
\usepackage[nocompress]{cite}
\else
\usepackage{cite}
\fi

\definecolor{Gray}{gray}{0.92}
\newcolumntype{u}{>{\columncolor{Gray}}c}

\newtheorem{lemma}{Lemma}[section]

\newtheorem{corollary}{Corollary}[section]
\newtheorem{definition}{Definition}[section]

\newcommand*{\horzbar}{\rule[.6ex]{2.5ex}{0.5pt}}

\newcommand{\transpose}{^\mathsf{T}}

\def\diamG{\text{diam}_{\mathcal{G}}}

\def\bX{\mathbf{X}}
\def\bS{\mathbf{S}}

\def\bG{\mathbf{G}}
\def\bD{\mathbf{D}}

\def\bN{\mathbf{N}}

\def\bU{\mathbf{U}}

\def\bQ{\mathbf{Q}}
\def\bK{\mathbf{K}}

\def\bXT{\mathbf{X}\transpose}

\def\cG{\mathcal{G}}

\def\cC{\mathcal{C}}
\def\cQ{\mathcal{Q}}
\def\cD{\mathcal{D}}

\def\cV{\mathcal{V}}

\def\cP{\mathcal{P}}

\def\bx{\mathbf{x}}

\def\MSE{\text{MSE}}

\def\reals{\mathbb{R}}

\def\bPhi{\mathbf{\Phi}}

\def\beta{\mathbf{\eta}}
\def\bPhi{\mathbf{\Phi}}
\def\bPhif{\mathbf{\Phi}_{\mathsf{FG}}}
\def\bPhir{\mathbf{\Phi}_{\mathsf{RP}}}
\def\bPhin{\mathbf{\Phi}_{\mathsf{Nys}}}

\DeclareMathOperator{\Var}{Var}

\begin{document}

\title{Recursive nearest agglomeration (ReNA): fast clustering for
approximation of structured signals}

\author{~Andr\'es~HOYOS-IDROBO, ~Ga\"el~VAROQUAUX, ~Jonas~KAHN,  
and~Bertrand~THIRION
\IEEEcompsocitemizethanks{\IEEEcompsocthanksitem
A. Hoyos-Idrobo, G.Varoquaux and B.Thirion are with Parietal team, Inria, 
CEA, University Paris-Saclay,
91191, Gif sur Yvette, France.\protect\\
E-mails: firstname.lastname@inria.fr}
\IEEEcompsocitemizethanks{\IEEEcompsocthanksitem
J. Kahn is with the 
Institut de Math\'ematiques de Toulouse, UMR5219, Universit\'e de
Toulouse, CNRS, France\protect.
E-mail: jonas.kahn@math.univ-toulouse.fr}
}

\markboth{IEEE Transactions on Pattern Analysis and Machine   
Intelligence,~Vol.~XX, No.~XX, 2016}%
{Hoyos-Idrobo \MakeLowercase{\textit{et al.}}: 
Fast approximation of structured signals with clustering: Recursive nearest 
neighbor (ReNA)}

\IEEEcompsoctitleabstractindextext{%
\begin{abstract}
In this work, we revisit fast dimension reduction approaches, as with random 
projections and random sampling.
Our goal is to summarize the data to decrease computational costs and
memory footprint of subsequent analysis.
Such dimension reduction can be very efficient when the signals of
interest have a strong structure, such as with images. 
We focus on this setting and investigate feature clustering schemes for data
reductions that capture this structure. 
An impediment to fast dimension reduction is then that good clustering comes 
with large algorithmic costs.
We address it by contributing a linear-time agglomerative clustering scheme, 
Recursive Nearest Agglomeration (ReNA). Unlike existing fast agglomerative 
schemes, it avoids the creation of giant clusters.
We empirically validate that it approximates the data as well as traditional 
variance-minimizing clustering schemes that have a quadratic complexity.
In addition, we analyze signal approximation with feature clustering and
show that it can remove noise, improving subsequent analysis steps. 
As a consequence, data reduction by clustering features with ReNA
yields very fast and accurate models, enabling to process large 
datasets on budget.
Our theoretical analysis is backed by extensive experiments on
publicly-available data that illustrate the computation efficiency and
the denoising properties of the resulting dimension reduction scheme.
\end{abstract}

\begin{IEEEkeywords}
clustering, dimensionality reduction, matrix sketching, classification, 
neuroimaging, approximation
\end{IEEEkeywords}
}

\maketitle

\IEEEraisesectionheading{\section{Introduction }\label{sec:introduction}}

\IEEEPARstart{C}HEAP and ubiquitous sensors lead to a rapid increase
of data sizes, not only in the sample direction --the number of
measurements-- but also in the feature direction. 
\textit{Features} refer here to the dimensions of each observation: pixels/voxels of images, time points of signals, loci of genotypes etc.
These ``big data'' put a lot of strain on data management
and analysis.
Indeed, they entail large memory and storage footprint,
and the algorithmic cost of querying or processing them is often
super-linear in the data size.
Yet, such data often display a low-dimensional structure,
for instance originating from the physical process probed by the sensor.
Thus, the  data can be well approximated by
a lower-dimension representation, dropping drastically the cost
of subsequent data management or analysis. 
The present paper focuses on these fast signal approximations.\\

The deluge of huge sensor-based data is ubiquitous:
in imaging sciences --\emph{e.g.}
biological~\cite{Amat2013} or medical~\cite{VanEssen2012}--, 
genomics~\cite{Overbeek2000,Cole2013}, and seismology~\cite{addair2014seismic}, 
to name a few applications.
Taming the computational costs created by the rapid
increase in signal resolution is an active research question.
Many approaches integrate reduced signal representations in
statistical analysis~\cite{Amat2013, jegou2010improving, hoyos2015, 
Chakrabarti:2002:LAD:568518.568520,keogh2001dimensionality}.\\

In signal processing and machine learning, fast signal approximation is
central to speeding up algorithms: approximating kernels~\cite{rahimi2007}, 
fast approximate nearest neighbors~\cite{Ailon2009},
or randomized linear algebra~\cite{Halko2011}.
Note that for all these,  the only requirement on the reduced
representation is that it preserves pairwise distances between signals.
data reduction makes processing huge data sets much easier as it
decreases both memory requirements and computation time.
Indeed, even for linear complexity algorithms, the computation cost of
growing data size is worse than linear once data no longer fit in
cache or memory.

There are a wide variety of standard data-compression approaches.
Typically, a data matrix is represented by a sketch matrix that is
significantly smaller than the original, but approximates it well. 
Two sketching strategies are commonly employed:
\emph{i)} approximating the matrix by a small subset of
its rows (or columns) (e.g. Nystr\"om~\cite{GittensM13} and 
CUR~\cite{mahoney2009matrix});
\emph{ii)} randomly combining matrix rows, relying on subspace
embedding and strong concentration phenomena, e.g.
random projections~\cite{Johnson1984}.
Random projections are appealing as they come with
theoretical guaranties on the expected distortion.
For information retrieval, state-of-the-art indexing of times
series can be achieved with a symbolic representation~\cite{lin2007} that
finds a regular piecewise constant approximation of the signal.

Here, we are interested in representing \emph{structured} signals,
and using this structure to improve the data approximation and speed up
its computation.
Such signals are often modeled as generated from a random process acting on a 
topology.
Individual features of the data then form vertices of a graph.
Edges can be predefined by the specificity of the acquisition process,
such as the physics of the sensors.
Thus, the connectivity of the features is independent from
the data themselves. 
Henceforth,  we refer to this connectivity as
\emph{structure}\footnote{We consider the structure as prior information, and 
assume it static across time.}.
Note that such a description is not limited to regular grids, such
as time-series or images, and encompasses for instance data
on a folded surface~\cite{shuman2013emerging}.

Our contribution focuses on fast dimension reduction that adapts to
common statistics across the data, as with
the Nystr\"om approximation, but unlike random projections.
Our goal is to speed up statistical analysis, \emph{e.g.}
machine learning, benefiting from the
structure of the data.
For this, we use \emph{feature grouping},
approximating a signal by partitioning its feature space  
into several subsets and replacing the value of each subset 
with a constant value.
We use a clustering algorithm to adapt the partition to the
data statistics.
Agglomerative clustering algorithms are amongst the fastest
approaches to extract many clusters with graph-connectivity
constraints.
However, they fail to create clusters of evenly-distributed size,
favoring a few huge clusters\footnote{This phenomenon is
known as percolation in random graphs~\cite{stauffer1992}.}.

\paragraph{\sl Contributions} our contributions are
two-fold.

\emph{i)} We analyze dimensionality reduction of structured signals by
feature grouping.
We show that it has a denoising effect on these signals, hence
improves subsequent statistical analysis.
\emph{ii)} We introduce a fast agglomerative clustering that is well
suited to perform the feature grouping.
This clustering algorithm finds clusters of roughly even size in linear time,
maintaining meaningful information on the structure of the data.
Our pipeline is very beneficial for analysis of
large-scale structured datasets, as the dimension reduction is very fast,
and it reduces the computational cost of various statistical estimators
without losing accuracy.

The paper is organized as follows. In section 2, we give prior art on
fast dimension reduction and analyze the theoretical performance of feature
grouping. In section 3, we introduce ReNA, a new fast clustering
algorithm. In section 4, we carry out extensive
empirical studies 
comparing many fast dimension-reduction approaches and show on real-life
data that feature grouping can have a denoising effect.

\smallskip
\paragraph{\sl Notations}
Column vectors are written using bold lower-case, e.g., $\bx$. For a vector 
$\bx$, the $i$-th component of $\mathbf{x}$ is denoted $\bx_i$. 
Matrices are written using bold capital letters, e.g., $\bX$. 
The $j$th column vector of $\bX$ is denoted $\bX_{*, j}$, the $i$th row 
vector of $\bX$ is denoted $\bX_{i, *}$, and $[n]$ denotes $\{1, \ldots, 
n\}$.
Letters in calligraphic, e.g. $\cP$ denotes sets or graphs, and it will be 
clarified by the context.
Let $\{\cC_i\}_{i=1}^k$ be short for the set $\{\cC_1, \ldots, \cC_k\}$. 
$|\cdot|$ denotes the cardinality of a set.
The $\ell_p$ norm of a vector $\bx = [\bx_1, \bx_2, \ldots, \bx_k] \in 
\reals^k$ is 
defined as $\|\bx\|_p = \left(\sum_{i=1}^k |\bx_i|^p\right)^\frac{1}{p},$ for 
$p=[1, \infty)$. 
\section{\label{sec:dimension_reduction} Dimension reduction of structured 
signals}
In this section, we review useful prior art on random projections and random 
sampling.
Then we analyze signal approximation with feature grouping. 

\subsection{Background and related prior art} 
Signal approximation with 
random projections or random sampling techniques is now central to
many data analysis, machine learning, or signal processing algorithms.

Let $\bX \in \reals^{p\times n}$ be a data matrix composed of $n$ samples and 
$p$ features (i.e. pixels/voxels).
We are interested in an operator $\bPhi \in \reals^{k\times p}$ that reduces 
the dimension of the data in the feature direction, acting as a  preprocessing 
step to make further analysis more tractable. 
This operator should maintain approximately the pairwise distance between pairs 
of samples $(\bX_{*,i}, \bX_{*, j}) \in \bX^2$ for $(i,j) \in [n]^2$, 
\begin{equation}
\|\bPhi \ \bX_{*, i} - \bPhi \ \bX_{*, j}\|_2^2 \approx \|\bX_{*,i} 
- \bX_{*, j}\|_2^2, \ \ \forall (i, j) \in [n]^2. 
\label{eq:distortion_general}
\end{equation}
Note that this approximation needs to hold only on the data 
submanifold, and not the entire $\reals^p$.

\paragraph*{\textbf{Random projections}} A standard choice is to build $\bPhi$ 
with random projections, $\bPhir$~\cite{hedge2015}. 
It is particularly attractive due its algorithmic simplicity and  
theoretical guaranties that make it $\epsilon$-isometric (see 
Eq.~\ref{eq:distortion_rp}).
\begin{lemma}
By the Johnson-Lindenstrauss lemma~\cite{Johnson1984}, the pairwise 
distances among a collection $\mathcal{X}$ of $n$-points in $\mathbb{R}^p$ are 
approximately maintained when the points are mapped randomly to an Euclidean 
space of dimension $k = O(\epsilon^{-1}\log n)$ up to a distortion at most 
$\epsilon$.
More precisely, given $\epsilon, \delta \in (0, 1)$ and $k \leq p$, there 
exists a random linear projection $\bPhir : \reals^{p} \rightarrow \reals^{k}$ 
such that for every $\bx, \bx^\prime$ in $\mathcal{X}$, the following relations 
hold: 
\begin{equation}
(1 - \epsilon)\|\bx - \bx^\prime\|_2^2 \leq \|\bPhir \ \bx - \bPhir 
\ \bx^\prime\|_2^2 \leq (1 + \epsilon)\|\bx - \bx^\prime\|_2^2,
\label{eq:distortion_rp}
\end{equation}
with probability at least $1 - \delta$.
\label{lemma:random_projections}
\end{lemma}
Johnson-Lindenstrauss embeddings have been widely used in the last years.
By providing a low-dimensional representation of the data, 
they can speed up algorithms dramatically, in 
particular when runtime depends super-linearly on the data dimensionality.
In addition, as this representation of the data is accurate in the sense of 
the $\ell_2$ 
norm, it can be used to approximate shift-invariant 
kernels~\cite{LuDFU13,rahimi2007}.

The $\bPhir$ matrix can be generated by sampling from a Gaussian 
distribution with rescaling. 
In practice, a simple and efficient generation scheme can yield a very
sparse random matrix with good properties~\cite{achlioptas2003,tropp2011}.

This approach suffers from two important limitations: \emph{i)} 
inverting the random mapping from $\reals^{p}$ to $\reals^{k}$ is 
difficult, requiring more constraints on the data (e.g. sparsity), which
entails another estimation problem. As a result, it yields less meaningful or 
easily interpretable results, as the ensuing inference steps cannot be made 
explicit in the original space.
\emph{ii)} This approach is suboptimal for structured datasets, since it  
ignores the properties of the data, such as a possible spatial smoothness.

\paragraph*{\textbf{Random sampling}} A related technique is random 
sampling, and in particular the Nystr\"om approximation method. 
This method is mainly used to build a low-rank approximation of a matrix.
It is particularly useful with kernel-based methods when the number 
$n$ of samples is large, given that the complexity of building a kernel matrix 
is at least quadratic in $n$~\cite{Williams01usingthe}.
It has become a standard tool when dealing with large-scale 
datasets~\cite{GittensM13}.

The idea is to preserve the spectral structure of a kernel matrix $\bK$ using 
a subset of columns of this matrix, yielding a low-rank approximation.
This can be cast as building a data-driven feature mapping $\bPhin \in 
\reals^{k\times p}$. 
In a linear setting, the kernel matrix is defined as $\bK = \bXT \ \bX$, which  
leads to the following approximation:
\begin{equation}
\bK_{i, j} = \left\langle \bX_{*, i}, \bX_{*, j}\right\rangle 
\approx\left\langle \bPhin \ \bX_{*, i}, \bPhin \ \bX_{*, j}\right\rangle.
\end{equation}
Here, building a base $\bPhin$ is achieved by randomly sampling $k \ll
n$ points from $\bX$, and then normalizing them --i.e. obtaining an
orthogonal projector to the span of the subsampled data-- see
algorithm 2 in supplementary materials.
The cost of the SVD dominates the complexity of this method $O(p k\min\{p, 
k\})$.
This method is well suited for signals with a common structure, for
instance images that share a common 
spatial organization captured by $\bPhin$.
As the Nystr\"om approximation captures the structure of the data,
it can also act as a regularization~\cite{Rudi2015}.

\subsection{Dimension reduction by feature grouping}
Here we analyze feature grouping for signal approximation.

\subsubsection{The feature-grouping matrix and approximation}

\begin{figure}[b]
\centering
\includegraphics[width=1.\linewidth]{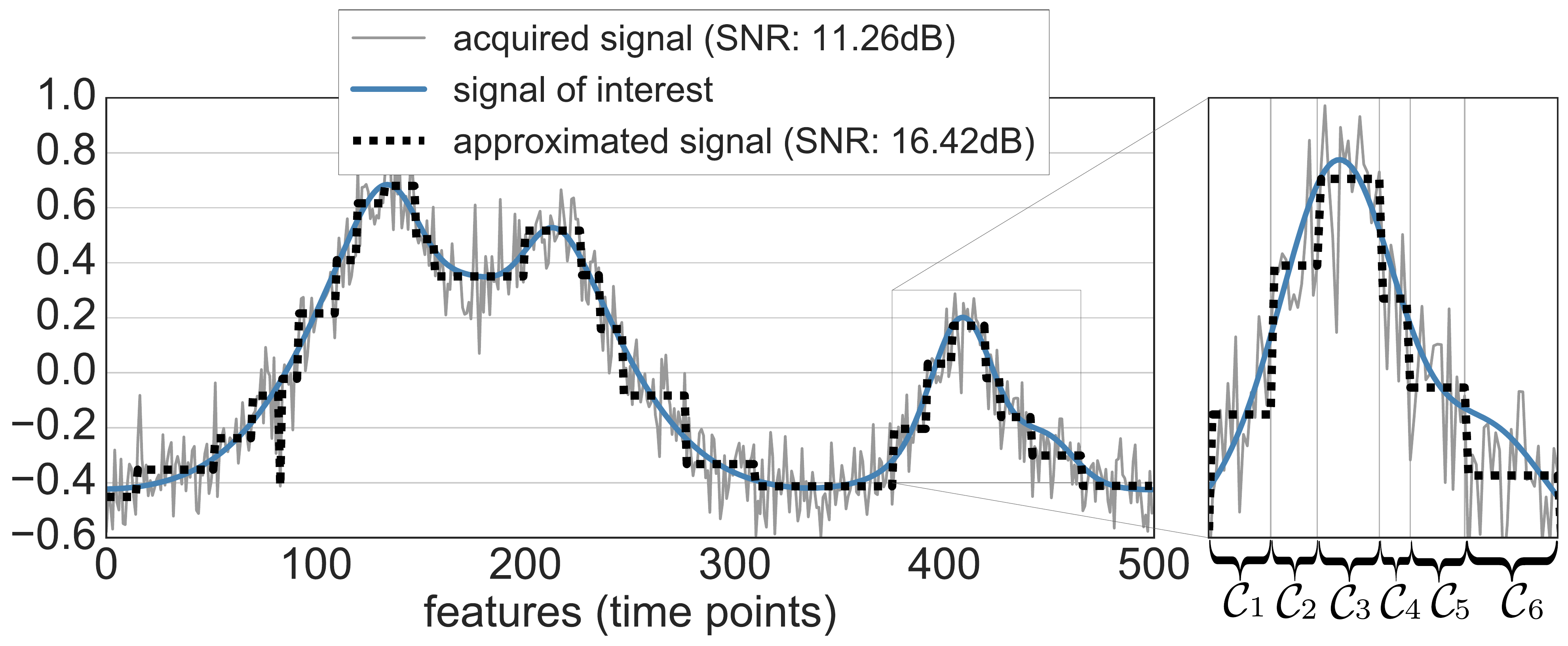}
\caption{\textbf{Illustration of the approximation of a signal:} Piece-wise 
constant approximation of a 1D signal contaminated with additive Gaussian 
noise, $\bx \in \reals^{500}$. This signal is a sample of our
statistical problem.
The approximation is built by clustering features (here time-points) with 
a connectivity constraint (i.e. using a spatially-constrained Ward clustering).
Only $25$ clusters can preserve the structure of the 
signal and decrease the noise, as seen from the 
signal-to-noise-ratio ($dB$).}
\label{fig:example_noise}
\end{figure}

Feature grouping defines a matrix $\bPhi$ that extracts piece-wise
constant approximations of the data~\cite{buhlmann2012}. 
Let $\bPhif$ be a matrix composed with constant amplitude groups (clusters).
Formally, the set of $k$ clusters is given by
$\cP = \{\cC_1, \cC_2, \ldots, \cC_k\}$, where each cluster $\cC_q \subset [p]$ 
contains a set of indexes that does not overlap other clusters, $\cC_q \cap 
\cC_l = \emptyset$, for all $q\neq l$.
Thus, $(\bPhif \ \bx)_q = \alpha_q \sum_{j \in \cC_q} \bx_j$  yields a 
reduction of a data sample $\bx$ on the $q$-th cluster, where $\alpha_q$ is a 
constant for each cluster.  
With an appropriate permutation of the indexes of the 
data $\bx$, the matrix $\bPhif$ can be written as
\begin{equation*}
\bPhif =
\left[
\begin{matrix}
\alpha_1 \horzbar \alpha_1 & 0 \horzbar 0 &\ldots & 0 \horzbar 0\\
0 \horzbar 0  & \alpha_2  \horzbar \alpha_2 & \ldots & 0 \horzbar 0\\
 \vdots & \vdots & \ddots & \vdots  \\
0 \horzbar 0  & 0 \horzbar 0 & \ldots & \alpha_k  \horzbar \alpha_k\\
\end{matrix}
\right] \in \reals^{k \times p}.
\end{equation*}
We choose $\alpha_q =  1/\sqrt{|\cC_q|}$ to set the non-zero singular values 
of $\bPhif$ to $1$, making it an orthogonal projection.
\smallskip

We call $\bPhif \ \bx \in \reals^k$ the \emph{reduced} version of $\bx$ and 
$\bPhif\transpose \bPhif \ \bx \in \reals^p$ the \emph{approximation} of $\bx$.
Note that having an approximation of the data means that the ensuing inference 
steps can be made explicit in the original space. 
As the matrix $\bPhif$ is sparse, this approximation follows the same
principle as~\cite{le2015flexible}, speeding up computational time and
reducing memory storage.

Let $M(\bx)$ be the approximation error for a data $\bx$ given a feature 
grouping matrix $\bPhif$, 
\begin{equation}
\label{eq:intertia_matrix}
M(\bx) = \left\|\bx - \bPhif\transpose\bPhif \ \bx\right\|_2^2,
\end{equation}
this is often called inertia in the clustering literature. 
This corresponds to the sum of all the local errors (the approximation error 
for each cluster), $M(\bx) = \sum_{q=1}^{k} m_q$, where $m_q(\bx)$ is the sum 
of squared differences between the values in the $q$-th cluster and its 
representative center, as follows
\begin{equation}
\label{eq:inertia_component}
m_q(\bx) = \left\| \bx_{\cC_q} - \frac{(\bPhif \ \bx)_{q}} 
{\sqrt{|\cC_q|}}\right\|_2^2,
\end{equation}
where $\bx_{\cC_q} \in \reals^{|\cC_q|}$ are the values $\bx_i$ such that $i 
\in \cC_q$.
The squared norm of the data $\bx$ is then decomposed in two terms: fidelity 
and inertia, taking the form (see section 2 in supplementary materials): 
\begin{equation}
\|\bx\|_2^2 = \underbrace{\|\bPhif \ \bx\|_2^2}_{\text{Reduced norm}} + 
\underbrace{\sum_{q=1}^{k} \left\| \bx_{\cC_q} - 
\frac{(\bPhif \ \bx)_q}{\sqrt{|\cC_q|}}\right\|_2^2}_{M(\bx)\text{: Inertia}}.
\label{eq:decomposition}
\end{equation}
Eq.~\ref{eq:decomposition} is key to understanding the desired
properties of a matrix $\bPhif$.
In particular, it shows that it is beneficial to work in a large $k$ regime to 
reduce the inertia.

\subsubsection{Capturing signal structure}

We consider data with a specific structure, e.g. spatial data.
Well-suited dimensionality reduction can leverage this structure to bound the
approximation error. 
We assume that the data $\bx \in \reals^p$ are generated from a
process acting on a space with a neighborhood structure (topology).
To encode this structure, the data matrix $\bX$ is  associated with an
undirected graph $\cG$ with $p$ vertices $\cV = \{v_1, v_2, \ldots,
v_p\}$.  Each vertex of the graph corresponds to column index in the data
matrix $\bX$ and the presence of an edge means that these features are
connected.
For instance, for 2D or 3D image data, the graph is a 2D or 3D 
lattice connecting neighboring pixels.
The graph defines a graph distance between features $\text{dist}_\cG$.
In practice, we perform the calculations with the adjacency matrix $\bG$ of the 
graph $\cG$.
\begin{definition} \textbf{$L$-Smoothness of the signal:}
\label{def:smoothness}
A signal $\bx \in \reals^p$ structured by a graph $\cG$, is pairwise Lipschitz 
smooth with parameter $L$ when it satisfies
\begin{equation}
|\bx_i - \bx_j| \leq L\, \text{dist}_\cG(v_i,\, v_j), \quad \forall (i, 
j) \in [p]^2.
\end{equation}
This definition means that the signal is smooth with respect to the 
graph that encodes the underlying structure.
Note that $\text{dist}_\cG$ has no unit since the scale is fixed by having each 
edge have length $1$.
\end{definition}
\begin{lemma}
\label{lemma:clustering_distortion}
Let $\bx \in \reals^p$ be a pairwise L-Lipschitz signal, and 
$\bPhif \in \reals^{k\times p}$ be a fixed feature grouping matrix, formed by 
$\{\cC_1, \ldots, \cC_k\}$ clusters. 
Then the following holds:
\begin{equation}
\|\bx\|_2^2 - L^2 \sum_{q=1}^{k}|\cC_q|\,\diamG(\cC_q)^2 \leq \|\bPhif \ 
\bx\|_2^2 \leq \|\bx\|_2^2,
\label{eq:distortion_partition_final}
\end{equation}
where $\diamG(\cC_q) = \sup\limits_{v_i, v_j \in \cC_q}\text{dist}_\cG(v_i,\, 
v_j)$.
\end{lemma}
See section 2 in supplementary materials 
for a proof. 

We see that the approximation is better if:

\emph{i) The cluster sizes are about the same.}
Even if the clusters are balls of dimension $d$, the $\diamG(\cC_q)$ is of 
order $|\cC_q|^{1/d}$. 
We thus expect the clusters to be compact. 

\emph{ii) The clusters have a small diameter.}
As the left-hand side of Eq.\ref{eq:distortion_partition_final} is upper 
bounded by $ L^2 \,\diamG(\cC_q)^2$, and $\diamG(\cC_q) \leq |\cC_q|$. 
We see that the clusters have to be small.

These arguments are based only on the assumption of smoothness of the signal.
Refining Eq.~\ref{eq:distortion_partition_final} gives an intuition on how
a partition could be adapted to the data:  
\begin{corollary}
\label{coro:distortion_refinement}
Let $L_q$ be the smoothness index inside cluster $C_q$, for all $q \in [k]$. 
This is the minimum $L_q$ such that:
\begin{equation*}
|\bx_i - \bx_j| \leq L_q \, \text{dist}_\cG(v_i,\, v_j), \quad \forall (i, j) 
\in C_q^2.
\end{equation*}
Then the following two inequalities hold:
\begin{equation}
\begin{split}
\|\bx\|_2^2 - \sum_{q=1}^{k}|\cC_q| \sup_{\bx_i, \bx_j \in \bx_{\cC_q}} |\bx_i 
- \bx_j|_2^2 \leq & \\
\|\bx\|_2^2 - \sum_{q=1}^{k}L_q^2\,|\cC_q|\,\diamG(\cC_q)^2
\leq  & \|\bPhif \ \bx\|_2^2. 
\end{split}
\end{equation}
\end{corollary}
We can see that the approximation is better if:
\emph{i)} the signal in a cluster is homogeneous (low $L_q)$;
\emph{ii)} clusters in irregular areas (high $L_q$) are smaller.

Clusters $\mathcal{P}$ of the graph $\cG$ are defined as  
connected components of a subgraph. 
In this context, the size of the largest group, $\max_{q\in [k]}|\cC_q|$, can 
be studied with percolation theory, that characterizes the appearance of a 
giant connect component as edges are added~\cite{stauffer1992}.
\subsubsection{Approximating signals with unstructured noise}
Noise hinders subsequent statistical estimation.
Feature grouping can reduce the noise under certain conditions.
To explore them, we consider an additive noise model: the acquired data $\bX$ 
is a spatially-structured signal of interest $\bS$ contaminated by 
unstructured noise $\bN$, 
\begin{equation}
\bX_{*, i} = \bS_{*, i} + \bN_{*, i}, \quad \forall i \in [n].
\end{equation}
Applying the feature grouping matrix $\bPhif$ to the acquired signal
reduces the noise via within-cluster averaging.
In particular, i.i.d. noise with zero-mean and variance $\sigma^2$, leads
to the following relation between the Mean Squared Error of the approximated 
data, 
$\MSE_{\text{approx}}$, and the non-reduced one $\MSE_{\text{orig}}$ (see 
section 3 in supplementary materials):  
\begin{equation}
\MSE_{\text{approx}} \leq L^2 \sum_{q=1}^{k}|\cC_q|\,\diamG(\cC_q)^2 + 
\frac{k}{p}\, \MSE_{\text{orig}},
\label{eq:feature_grouping_denoising}
\end{equation}
where $p$ denotes the number of features (i.e. pixels/voxels), and $k$ denotes 
the number of clusters.
This gives a denoising effect if the smoothness parameter satisfies 
\begin{equation}
\label{eq:smoothness}
L^2 \leq \frac{(p - k)}{\sum_{q=1}^{k}|\cC_q|\,\diamG(\cC_q)^2}\, \sigma^2.
\end{equation}
When the signal of interest is smooth enough and the cluster sizes are roughly 
even, the feature grouping will reduce the noise, preserving the information of 
the low-frequency signal $\bS_i$.
Fig.~\ref{fig:example_noise} presents a
graphical illustration of the reduction and denoising capabilities of feature 
grouping.

The challenge is then to define a good $\bPhif$, given that data-unaware 
feature partitions are sub-optimal, as they do not respect the underlying 
structures and lead to signal loss.
\section{\label{sec:ReNA} ReNA: A fast structured clustering algorithm}
\begin{figure}[htb]
\centering
\includegraphics[width=.8\linewidth]{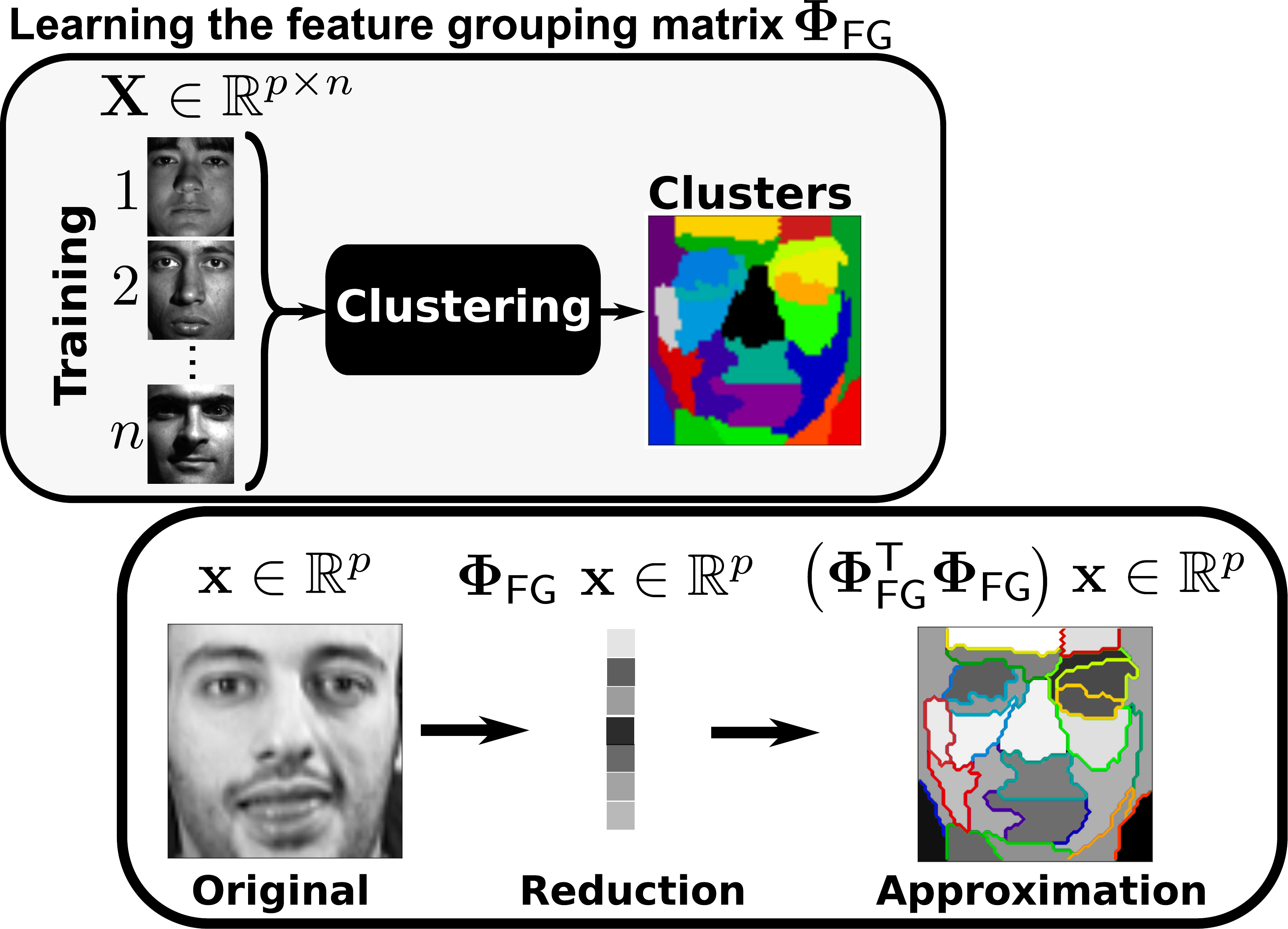} 
\caption{\textbf{Illustration of feature grouping:} In images,
feature-grouping data approximation corresponds to a super-pixel approach.
In a more general setting, this approach consists in finding a data-driven 
reduction $\bPhif$ using clustering of features. 
Then, the data $\bx$ are reduced to $\bPhif \ \bx$ and then used for further 
statistical analysis (e.g. classification). 
}
\label{fig:feature_grouping}
\end{figure}

In the feature-grouping setting above, we now consider a data-driven approach 
to build the matrix $\bPhif$.
We rely on feature clustering: a clustering algorithm is used to define
the groups of features from the data.
$\bX\in \reals^{p\times n}$ is represented by a 
reduced version $\bPhif \ \bX$, where $p$ is potentially very large (greater 
than $100\,000$), whereas $k$ is smaller but close enough to $p$ (e.g. $k = 
\lfloor p / 20 \rfloor$).
As illustrated in Fig.~\ref{fig:feature_grouping}, once the reduction
operator $\bPhif$ has been learned, it can be applied to new data coming 
form the same generative process.

\subsection{Existing fast clustering algorithms}
K-means clustering is a natural choice as it minimizes the total
inertia in Eq.~\ref{eq:decomposition}. 
But it tends to be expensive in our setting: The conventional k-means 
algorithm has a complexity of $O(p\,k)$ per iterations~\cite{Achanta2012}. 
However, the larger the number of clusters, the
more iterations are needed to converge, and the worst case complexity
is given\footnote{Note that
here $n$ and $p$ are swapped compared to common clustering literature, as
we are doing \emph{feature} clustering.} by 
$O(p^{k+2/n})$~\cite{arthur2006kmeans}.
This complexity becomes prohibitive with many clusters.

\paragraph*{\bf Super-pixel approaches}
In computer vision, feature clustering can be related to the notion of
\emph{super-pixels} (super-voxels for 3D images). The most common fast
algorithm for super-pixels is SLIC~\cite{Achanta2012},
which has a low computational cost and produces super-pixels/super-voxels of 
roughly even sizes. 
SLIC performs a local clustering of the image values with a spatial constraint, 
using as a distance measure the combination of two Euclidean distances: image 
values and spatial positions.
The SLIC algorithm is related to K-means, but it performs a fixed small
number of iterations, resulting in a complexity of $O(np)$. 
In the large-$k$ regime, it can be difficult to control precisely
the number of clusters, as some clusters often end up empty in the final 
assignment.

\paragraph*{\bf Agglomerative clustering}
Agglomerative clustering algorithms are fast in the setting of a large 
number $k$ of clusters. Unlike most clustering algorithms, adding a
graph structure constraint makes them even faster, as they can then discard 
associations between non-connected nodes.

Agglomerative clustering schemes start off by placing every data
element in its own cluster, then they proceed by merging repeatedly
the closest pair of connected clusters until finding the desired
number of clusters~\cite{hastie09statisticallearning}.
Various methods share the same approach, differing only in the linkage 
criterion used to identify the clusters to be merged. The most common 
linkages  are single, average, complete~\cite{hastie09statisticallearning} and 
Ward~\cite{ward1963}.
Average-linkage, complete-linkage, and Ward are generally preferable over 
single-linkage, as they tend to yield more balanced clusters. %
Yet single-linkage clustering is often used as it is markedly faster;
it can be obtained via a Minimum Spanning Tree and has a complexity of $O(np + 
p\log p)$~\cite{Mullner2011}.
Single linkage is also related to finding
connected components of a similarity graph \cite{mirkin1998mathematical},
an approach often used to group pixels on images
\cite{rosenfeld1966sequential}.
Average-linkage, complete-linkage and Ward are more costly, as they have a worst case complexity of 
$O(np^2)$~\cite{Mullner2011}.

The approximation properties of feature grouping are given by the
distribution of cluster sizes and the smoothness of the signal 
(Eq.~\ref{eq:distortion_partition_final}). Balanced
clusters are preferable for low errors.
Nevertheless, agglomerative clustering on noisy data often leads to a
``preferential attachment'' behavior: large clusters grow faster
than smaller ones. In this case, the largest cluster dominates the distortion, 
as in Lemma.~\ref{lemma:clustering_distortion}.
By considering the clusters as connected components on a similarity
graph, this behavior can be linked to percolation theory~\cite{stauffer1992},
that characterizes the appearance of a giant connected component (i.e. 
a huge cluster). In this case, the clustering algorithm is said to 
\textit{percolate}, and thus cannot yield balanced cluster sizes.

In brief, single-linkage clustering is fast but suffers form percolation 
issues~\cite{Penrose1995} and Ward's algorithm performs often well
in terms of goodness of fit for large $k$~\cite{thirion2014}.

More sophisticated agglomerative strategies have been proposed in the framework 
of computer vision (e.g.~\cite{Felzenszwalb2004}), but they have not been 
designed to avoid percolation and do not make it possible to control the number 
$k$ of clusters.
\subsection{Contributed clustering algorithm: ReNA}
For feature clustering on structured signals, an algorithm should take 
advantage of the generative nature of the data, e.g. for images,
work with local image statistics. 
Hence we rely on neighborhood graphs~\cite{Eppstein1997}.

Neighborhood graphs form an important class of geometric graphs with many 
applications in signal processing, pattern recognition, or data clustering. 
They are used to model local relationships between data points, with 
$\epsilon$-neighborhood
graphs or k-nearest neighbor graphs.
The $\epsilon$-nearest neighbor graph is the core of clustering methods for 
which the number $k$ of clusters is implicitly set by the $\epsilon$ 
neighborhood's radius~\cite{Ester96adensity-based}.
$K$-nearest neighbor graphs are also used for clustering and theoretical
results show that they can identify high-density modes of the
samples~\cite{MAIER20091749}.
However, measurement noise on between-sample similarities hinders their
recovery of this structure as neighborhood graphs tend to percolate
\footnote{\cite{MAIER20091749} explicitly used structured random graphs in
their analysis and exclude Erd\"os-R\'enyi graphs, \emph{i.e.} random
connections, as created by unstructured noise.}
for $k$ greater than or equal to $2$.
In contrast, the 1-nearest neighbor graph (1-NN) is not likely to 
percolate~\cite{Teng2007}. 
For this reason, we use the 1-NN graph.

In a nutshell, our algorithm relies on extracting the connected components
of a 1-NN graph.
To reach the desired number $k$ of clusters, we apply it recursively.
The algorithm outline is as follows:

\begin{description}
\item[\bf Initialization:]
We start by placing each of the $p$ features of the data $\bX$ in its own 
cluster 
$\cP = \{\cC_1, \ldots, \cC_p\}$.
We use the binary adjacency matrix $\bG \in \{0, 1\}^{p\times p}$ of the graph 
$\cG$, that encodes the topological structure of the features (i.e. one values 
denote connected vertices, whereas zero represents non-connected vertices).
\item[\bf Nearest neighbor grouping:]
We build the similarity graph which encodes the affinity between features.
We then find the nearest neighbor graph of this similarity graph, and
extract the connected components of this subgraph to reduce the data 
matrix $\bX$ and the topological structure $\bG$. 
Hence, we use the 1-nearest neighbor of each vertex (i.e. feature) of the 
similarity graph as linkage criterion.
These operations are summarized in the next steps:

\begin{enumerate}
\item \textbf{Graph representation:}
We build the similarity graph $\cD$ of the data $\bX$, represented by the 
adjacency matrix $\bD \in \reals^{p\times p}$.
The weights in $\bD$ are only assigned for edges in $\bG$
\footnote{This corresponds to an element-wise condition, where a similarity 
weight is assigned only if the edges 
are connected according to $\bG$.}.
\item \textbf{Finding 1-NN:}
Creating a 1-nearest neighbor graph $\cQ$, represented by the 
matrix $\bQ \in \{0, 1\}^{p\times p}$, where each vertex of $\cD$ is 
associated with its 
nearest neighbor in the sense of the dissimilarity measure (e.g. Euclidean 
distance, although a distance is not needed).
\item \textbf{Getting the clusters:}
We use~\cite{Pearce05animproved}\footnote{This algorithm is implemented in the 
SciPy package. However, other variants of \cite{tarjan1972depth} can be used.} 
to extract the set of connected components of 
$\bQ$ and assign them to the new set of clusters $\cP$.
\item \textbf{Reduction step:}
The clusters are used to reduce the graph $\cG$ and the data $\bX$. 
This boils down to averaging features and grouping edges.
\end{enumerate}
\item[\bf Stopping condition:] Nearest neighbor grouping can be 
performed repeatedly on the reduced versions of the graph $\cG$ and the 
data $\bX$ until the desired number $k$ of clusters is reached.
\end{description}
\begin{algorithm} 
\small
\caption{Recursive nearest neighbor (ReNA) clustering}
\label{al:ReNA}
\begin{algorithmic}[1]
\REQUIRE Data $\bX \in \reals^{p\times n}$, sparse
matrix $\bG \in \reals^{p\times p}$ representing the associated
connectivity graph structure, nearest-neighbor subgraph extraction function 
$\text{NN}$, connected components extraction function $\text{ConnectComp}$ 
\cite{Pearce05animproved}, 
desired number $k$ of clusters.
\ENSURE Clustering of the features $\cP = \{\cC_1, \cC_2, \ldots, \cC_k\}$ 
\STATE $q = p$ \COMMENT{Initializing the number of clusters to $p$}
\STATE $t = 0$
\STATE $\bX^{(t)} = \bX$\\
\STATE $\bG^{(t)} = \bG$
\WHILE{$q > k$}
    \STATE $\bD_{i,j}^{(t)} \leftarrow \begin{cases}
    \|\bX_{i, *}^{(t)} - \bX_{j, *}^{(t)}\|_2^2 & \text{if } \bG_{i, j}^{(t)} 
    \neq 0\\
    \infty & \text{otherwise}
    \end{cases}, (i,j) \in [q]\times[|\cP|]$\\    
    \COMMENT{Create a similarity weighted graph.}
    \STATE $\bQ\leftarrow \text{NN}(\bD^{(t)})$ \COMMENT{1-nearest neighbor 
    graph.}\\ 
    \STATE $\cP \leftarrow \text{ConnectComp}(\bQ),$ \\ 
    \COMMENT{Sets of connected components of 1-nearest neighbor graph.} 
    \STATE $\bU_{i, j} \leftarrow \begin{cases}
    1 & \text{if } i\in\cC_j\\
    0 & \text{otherwise}
    \end{cases}, (i,j) \in [q]\times[|\cP|]$\\
    \COMMENT{Assignment matrix}
    \STATE $\bX^{(t + 1)} \leftarrow 
    (\bU\transpose\bU)^{-1}\bU\transpose\bX^{(t)}, \ \bX^{(t + 1)} \in 
    \reals^{|\cP|\times n}$\\
    \COMMENT{Reduced data matrix. Note that the computation boils down to
     averaging grouped features.}
    \STATE $\bG^{(t + 1)} \leftarrow \text{support}(\bU\transpose \bG^{(t)} 
    \bU), \ \bG^{(t + 1)} \in \reals^{|\cP|\times|\cP|}$\\
    \COMMENT{Reduced between-cluster topological model; the non-zero values are 
    then replaced by ones.}
    \STATE $q = |\cP|$\COMMENT{Update the number of clusters}\\
    \STATE $t = t + 1 $\\
    \ENDWHILE
\RETURN $\cP$
\end{algorithmic}
\end{algorithm}

The algorithm is iterated until the desired number of clusters 
$k$ is reached.
At each iteration, a connected components routine extracts them from $\bQ$ and 
returns them as a set of clusters $\cP$. 
In the last iteration of the algorithm, if there are less than $k$ connected  
components, $\bQ$ is pruned of its edges with largest edge values to  keep 
only the $q-k$ shortest edges, so that no less than $k$ components are formed.

The number of iterations is at most $O\{\log(p/k)\}$ as the number of vertices 
is divided by $2$ (at least) at each step; in practice, we never have to go beyond 
$5$ iterations.
The cost of computing similarities is linear in $n$ and, as
all the operations involved are also linear in the number of vertices $p$, the 
total procedure is $O(np)$.

Even in presence of noise, the cluster diameter does not grow fast
thanks to the small number of iterations, hence there is a denoising
behavior from Eq.~\ref{eq:feature_grouping_denoising}.

\begin{figure}
\centering
\includegraphics[width=1. 
\linewidth]{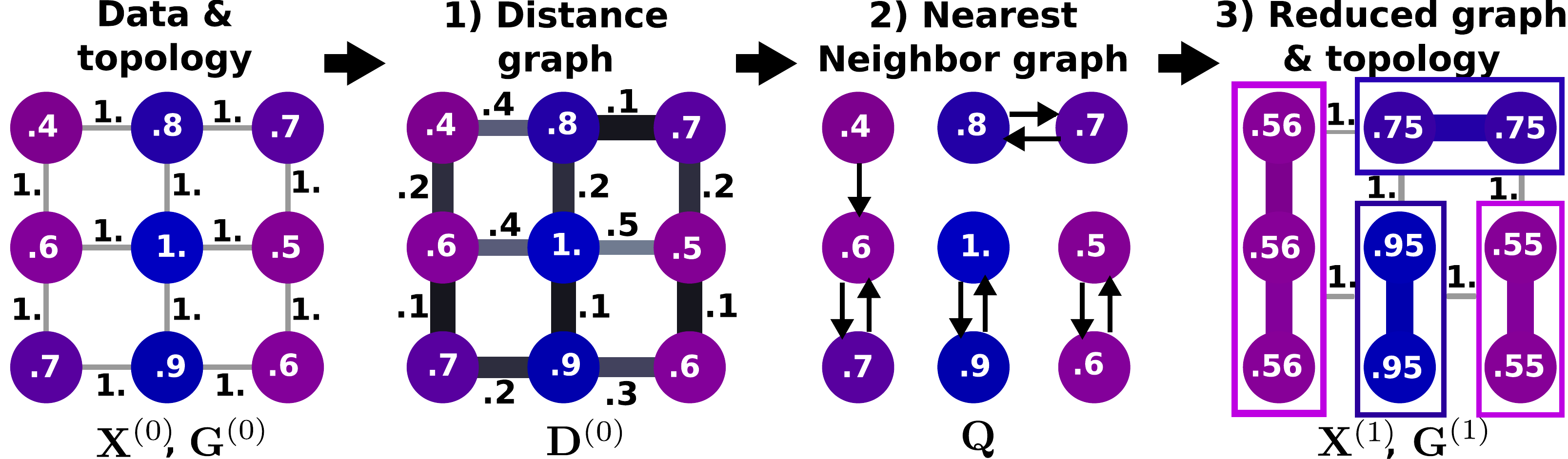}
\caption{\textbf{The nearest neighbor grouping:} 
The algorithm receives a data matrix $\bX$ represented on a regular square 
lattice $\bG$. \emph{left)} The nodes correspond to the feature values and the 
edges are the encoded topological structure.
\emph{1) Graph representation:} We calculate the similarity matrix $\bD$. 
\emph{2) Finding 1-NN:} We proceed by finding the 1-nearest neighbors subgraph 
$\bQ$ according to the similarity measure.
\emph{3) Getting the clusters and reduction step:}  We extract the connected 
components of $\bQ$ and merge the connected nodes.}
\label{fig:pedagogical_ReNA_iter}
\end{figure}

Note that the NN are calculated on the non-zero values encoded by $\bG$ 
(structure).
Additionally, the connected components are not symmetric (see 
Fig.~\ref{fig:pedagogical_ReNA_iter}). 
Thus, we simply take $\bQ = (\bQ + \bQ\transpose)$\footnote{This corresponds to 
a logical or operation.} 
to symmetrize them. 
We also use sparse matrices to perform all calculations hence we avoid 
forming $p\times p$ matrices.

\begin{figure*}
\begin{minipage}{.3\linewidth}
\caption{\textbf{Illustration of the working principle of the Recursive Nearest 
Neighbor, ReNA:} The white lines represent the edges of the connectivity graph. 
The algorithm receives the original sequence of images, considering each 
feature (i.e. pixel or voxel in the image) as a cluster.
From now on, for each iteration, the nearest clusters are merged 
(i.e. removing edges from the connectivity graph), 
yielding a reduced graph, until the desired number of clusters is found.}
\label{fig:pedagogical}
\end{minipage}\hfill%
\begin{minipage}{.69\linewidth}
    \includegraphics[width=1.\linewidth]{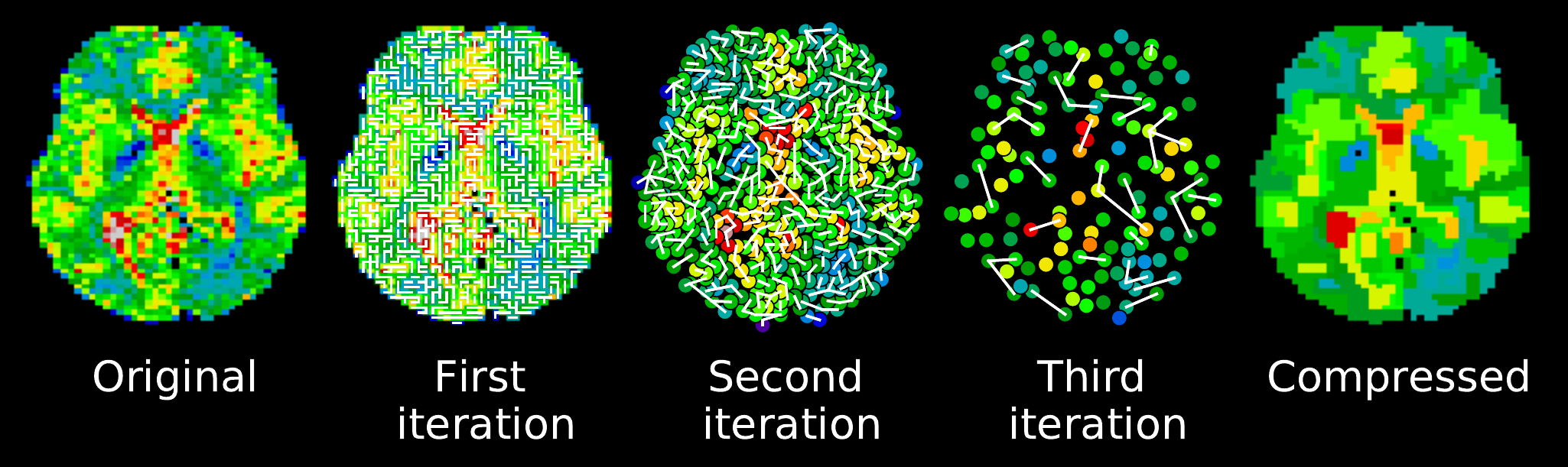}
\end{minipage}
\end{figure*}

Fig.~\ref{fig:pedagogical_ReNA_iter} presents one iteration of the nearest 
neighbor grouping on a regular square lattice. 
The pseudo-code of ReNA is given in algorithm~\ref{al:ReNA} and an illustration 
on a 2D brain image in Fig.~\ref{fig:pedagogical}.

\section{Experimental Study}
In this section, we conduct a series of experiments to assess the quality of 
the dimensionality reduction scheme and its viability as a preprocessing step 
for several statistical analyses.
Table.~\ref{tab:datasets} gives a summary of the datasets used.

We investigate feature grouping with a variety of
clustering algorithms: single-linkage, average-linkage, complete-linkage,
Ward, SLIC, and ReNA. 
We use the Euclidean distance for all algorithms and for all hierarchical
clustering methods we use the spatial structure as constraints on the
agglomeration steps (as in \cite{Felzenszwalb2004}). 
We compare them to other fast dimensionality reductions: random projections, 
random sampling, as well as image downsampling. 
We measure their ability to represent the data and characterize their 
percolation behavior when it is relevant.
We use prediction to evaluate their denoising properties.
To characterize beyond $\ell_2$ approximation, we also consider 
methods relying on higher moments of the data distribution: $\ell_1$ 
penalization and independent component analysis (ICA).
Note that downsampling images with linear interpolation can be seen 
as using data-independent clusters, all of the same size.

We present results as a function of the fraction of the signal, 
the ratio between the number $k$ of components and its largest possible value. 
We have two cases:
\emph{i)} for random projections and feature grouping
the ratio is $k/p \times 100\%$;
\emph{ii)} for random sampling
the ratio corresponds to $k/n \times 100\%$.

\subsection{Datasets}
\begin{table}
\caption{Summary of the datasets and the tasks performed with them.}%
\centering
\begin{adjustbox}{width=1\linewidth}
\sffamily
\begin{tabular}{  l  c  c c r }
\hline
Dataset & Description & $n$ & $p$ & Task \\
\hline
 &  & $10$ &  $\{8, 16, 64, 128\}^3$ & Time complexity (supp mat) \\ 
\multirow{-2}{*}{Synthetic}  & 
\multirow{-2}{*}{Cube} & $1\,000$ & $240\,000$ & Distortion\\
\cellcolor{Gray} & \cellcolor{Gray} Grayscale  & \cellcolor{Gray} & 
\cellcolor{Gray} & \cellcolor{Gray} Recognition of\\
\multirow{-2}{*}{\cellcolor{Gray} Faces~\cite{wright2009}} & \cellcolor{Gray} 
face images & 
\multirow{-2}{*}{\cellcolor{Gray} $2\,414$} & \multirow{-2}{*}{\cellcolor{Gray} 
$32\,256$} & \cellcolor{Gray} 38 subjects\\
 & Anatomical & & & Gender discrimination \\
\multirow{-2}{*}{OASIS~\cite{marcus2007}}  &   
brain images & \multirow{-2}{*}{$403$} & \multirow{-2}{*}{$140\,398$} & Age 
prediction\\
\cellcolor{Gray} & \cellcolor{Gray} Functional &\cellcolor{Gray} 
&\cellcolor{Gray} & \cellcolor{Gray}Predict 17 cognitive tasks\\
\multirow{-2}{*}{\cellcolor{Gray} HCP~\cite{VanEssen2012}} & \cellcolor{Gray} 
brain images & \multirow{-2}{*}{\cellcolor{Gray} $8\,294$} & 
\multirow{-2}{*}{\cellcolor{Gray} $254\,000$} & \cellcolor{Gray} Spatial ICA\\
\hline
\end{tabular}
\end{adjustbox}
\label{tab:datasets}
\end{table}
\subsubsection{Synthetic data} 
We generate a synthetic data set composed of $1\,000$ 3D images with and 
without noise. 
Each one is a cube of $p = {50}^3$ voxels containing a spatially smooth random 
signal (FWHM=$8$ voxels), our signal of interest $\bS$. 
The acquired signal $\bX$ is $\bS$ contaminated by 
zero-mean additive Gaussian noise, with a Signal-to-Noise Ratio (SNR) of 
$2.06dB$.

\begin{figure*}[bt!]
\centering
\subfigure[downsampling]{\includegraphics[width=.1445
\linewidth]{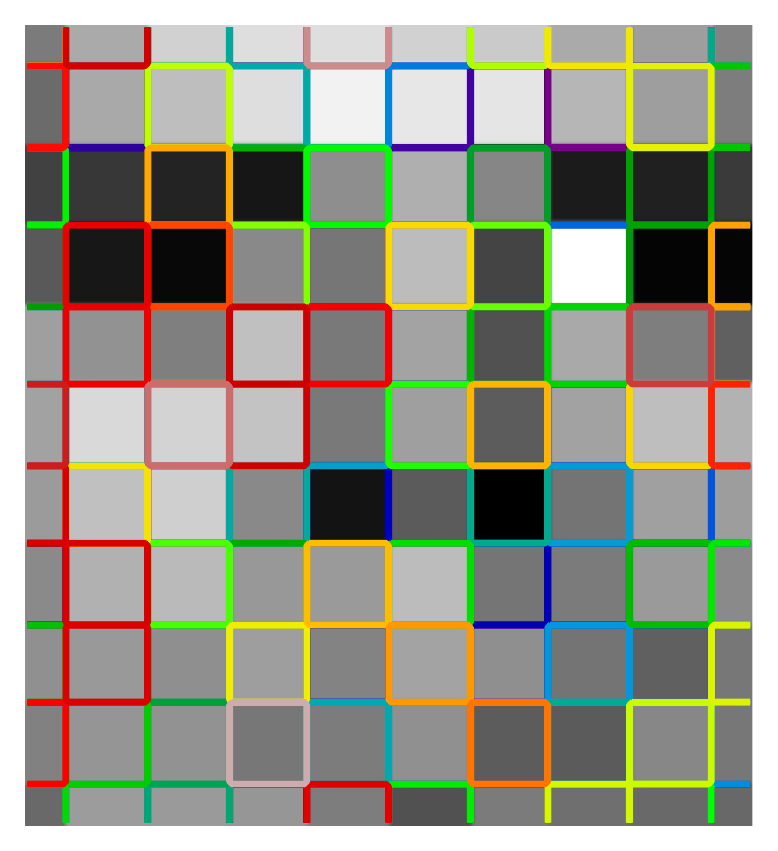}}\hspace{-3pt}
\subfigure[single-linkage]{\includegraphics[width=.1445
\linewidth]{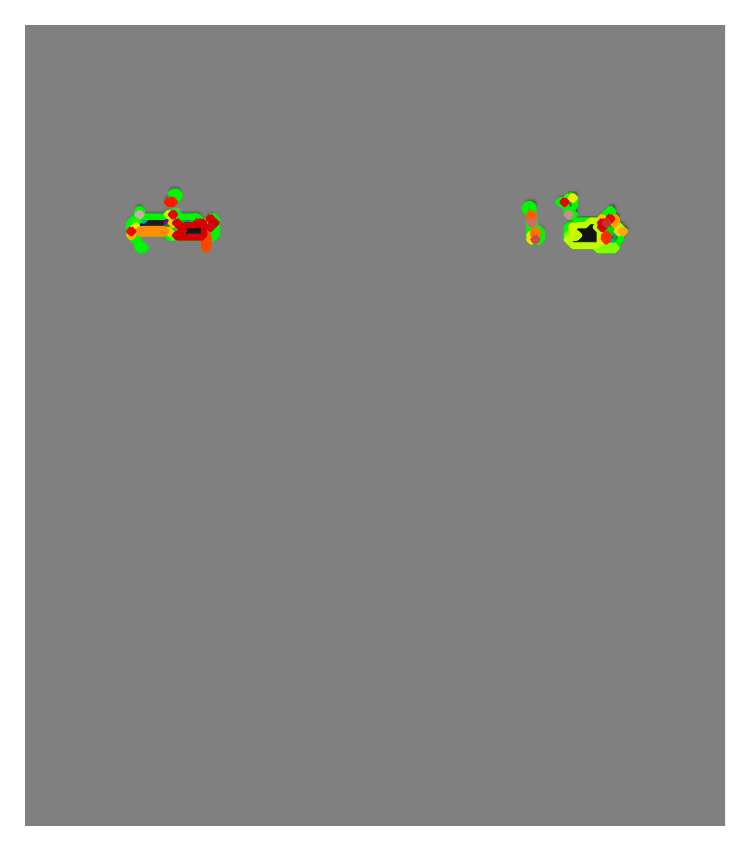}}\hspace{-3pt}
\subfigure[average-linkage]{\includegraphics[width=.1445
\linewidth]{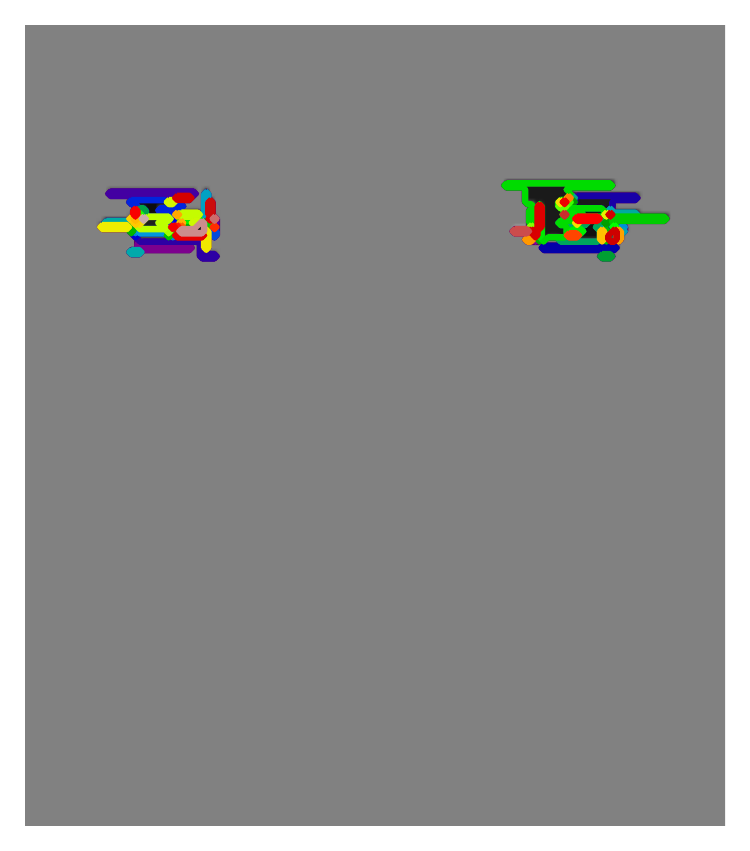}}\hspace{-3pt}
\subfigure[complete-linkage]{\includegraphics[width=.1445 
\linewidth]{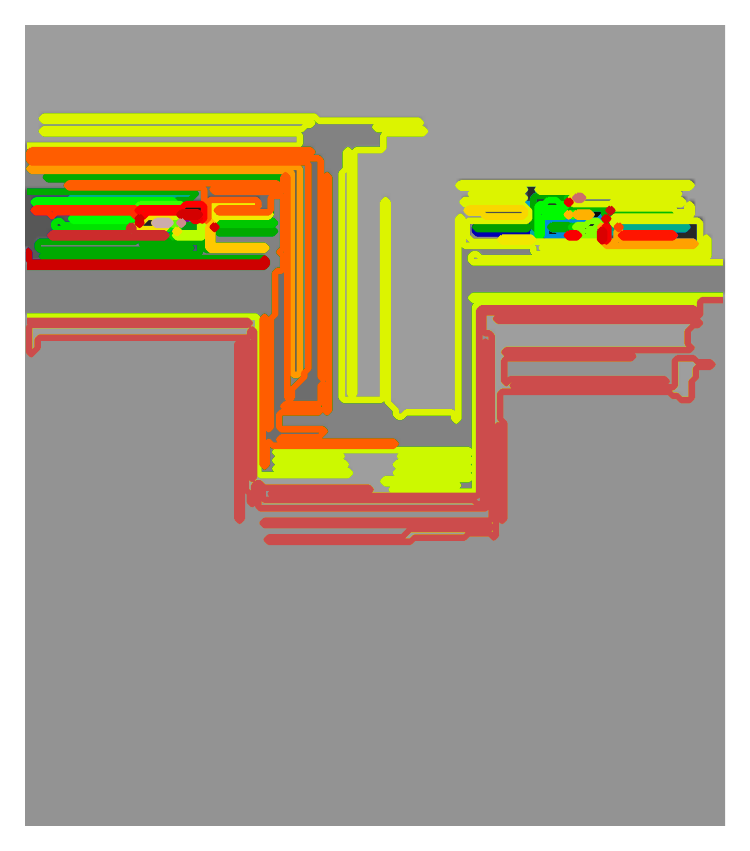}}\hspace{-3pt}
\subfigure[Ward]{\includegraphics[width=.1445
\linewidth]{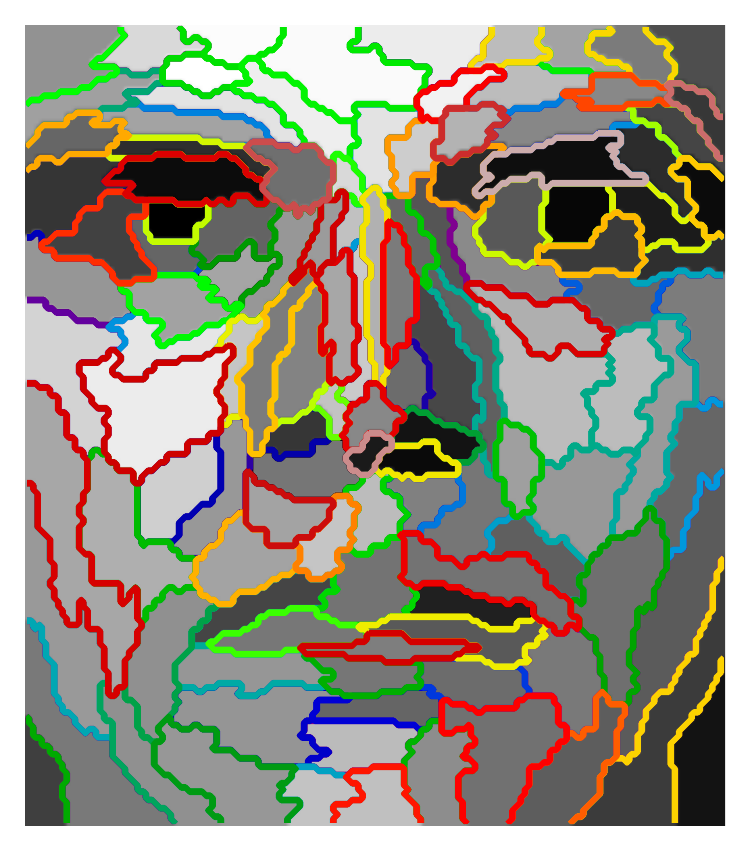}}\hspace{-3pt}
\subfigure[SLIC]{\includegraphics[width=.1445
\linewidth]{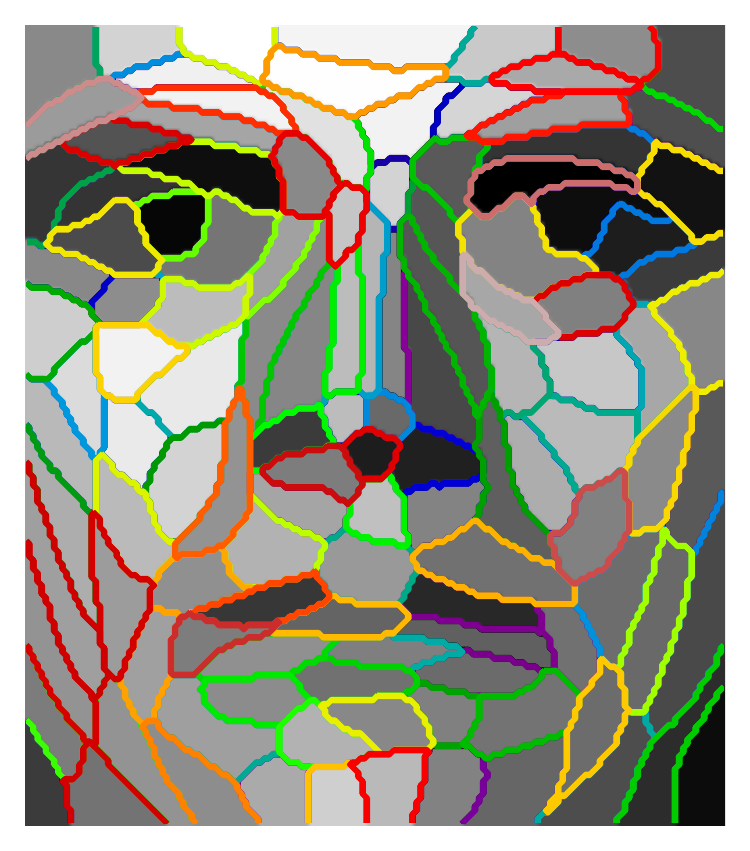}}\hspace{-3pt}
\subfigure[ReNA]{\includegraphics[width=.1445 
\linewidth]{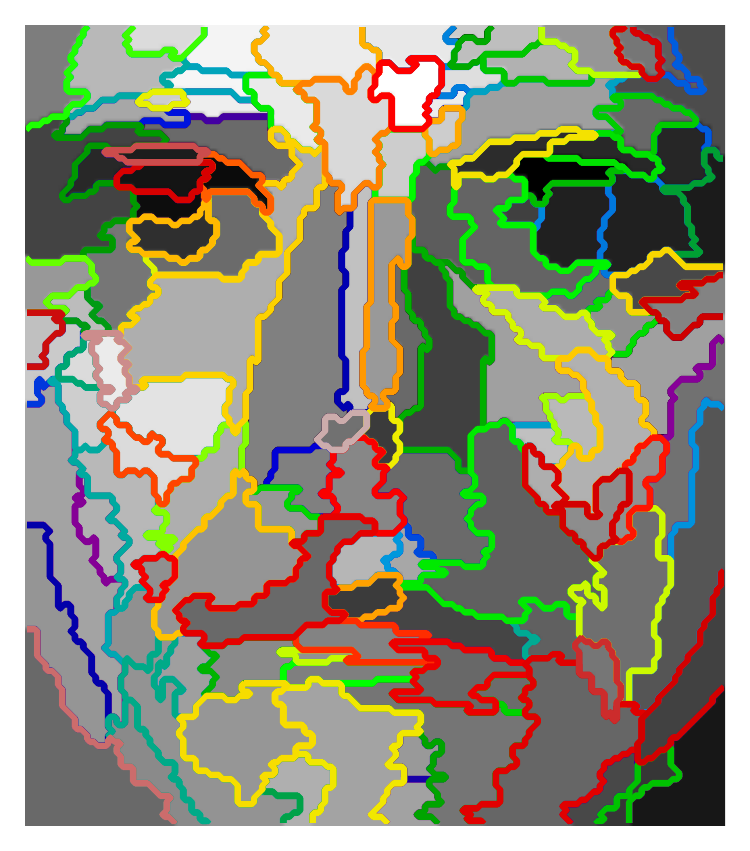}}\hspace{-3pt} 
\caption{
\textbf{Clusters obtained for the extended Yale B face dataset using 
various feature grouping schemes:} 
($k = 120$). 
Single, average and complete linkage clustering fail to represent the spatial 
structure of the data, finding a huge cluster leaving only small islands apart. 
Downsampling fails to capture the global appearance. 
In contrast, methods yielding balanced clusters maintain this structure.
Colors are random.}
\label{fig:example_clusters_faces}
\end{figure*}

\subsubsection{The extended Yale B face recognition dataset} 
This dataset was designed to study illumination effects on face 
recognition~\cite{wright2009} and consists of $n = 2\,414$ images of 
$38$ identified individuals  under $64$ lighting conditions. 
Each image was converted to grayscale, cropped, and normalized to $192 \times 
168$ pixels, leaving $p = 32\,256$ features. 
There are $38$ classes for the face recognition task, one per subject.
\subsubsection{The Open Access Series of Imaging Studies (OASIS)} 
The OASIS dataset\footnote{\small OASIS was 
supported by grants P50 AG05681, P01 AG03991, R01 AG021910, P50 MH071616, U24 
RR021382, R01 MH56584.}~\cite{marcus2007} consists of anatomical brain 
images (Voxel Based Morphometry) of $403$ subjects. 
These images were processed with the SPM8 software to obtain modulated grey 
matter density maps realigned to the Montreal Neurological Institute (MNI) 
template~\cite{Chau2005408} with a 2mm resolution.
Masking with a gray-matter mask 
yields $p = 140\,398$ voxels and $1$ GB of dense data.
We perform two prediction tasks: \emph{i)} Gender 
classification and \emph{ii)} age regression.

\subsubsection{Human Connectome Project (HCP)}
We use functional Magnetic Resonance Imaging (fMRI) data from 
the Human Connectome Project (HCP)~\cite{VanEssen2012}: 
500 participants (13 removed for quality reasons),
scanned at rest --typically analyzed via ICA-- and
during tasks~\cite{barch2013} --typically analyzed with linear
models.
We use the minimally preprocessed data~\cite{glass13}, resampled at
2mm resolution. 

\smallskip
\emph{Task data:} We use tasks relating to different cognitive
labels on working memory and cognitive control. 

\smallskip
\emph{Resting-state data:} We use the two resting-state sessions
from $93$ subjects. 
Each session represents data 
with $p\approx 220\,000$ and $n = 1\,200$, totaling $200\,\text{GB}$ of
dense data for all subjects and sessions.

\subsection{Technical Aspects}
We use scikit-learn for logistic 
and Ridge regression, fast-ICA, clustering, and random projections 
~\cite{Pedregosa2011}. 
We rely on scikit-image for SLIC~\cite{scikit-image}, 
on Nilearn to handle neuroimaging data, and 
on Scipy~\cite{jones2014scipy} to extract graph connected components.
Code for ReNA and experiments is available\footnote{\url{https://github.com/ahoyosid/ReNA}}.

\subsection{Quality assessment experiments}
We vary the number of clusters $k$ and
evaluate the performance of $\bPhi$ with three measures:
\emph{i)} the computation time; 
\emph{ii)} the signal distortion; 
\emph{iii)} the size of the largest cluster.
As a dimension reduction learned from data may capture noise in
addition to signal,  
we test the learned $\bPhi$ on left-out data, 
in a cross-validation scheme splitting the data randomly $50$ times. 
Each time, we learn $\bPhi$ on half of the noisy data and apply it
to the other half to measure distortion with regards to the non-noisy signal.
We vary the number of clusters $k \in [0.01 p, p]$. 
For Nystr\"om approximation, we vary the dimensionality $k \in [0.01 n, n]$.

Fig.~\ref{fig:example_clusters_faces} shows the clusters found by the 
various algorithms on the faces dataset. 
Single, average, and complete linkage have percolated, failing 
to retain the spatial structure of the faces. 
Downsampling also fails to capture this structure,  while Ward, SLIC and ReNA 
perform well in this task. 

\begin{figure}[bt]
\centering
\includegraphics[width=0.95\linewidth]{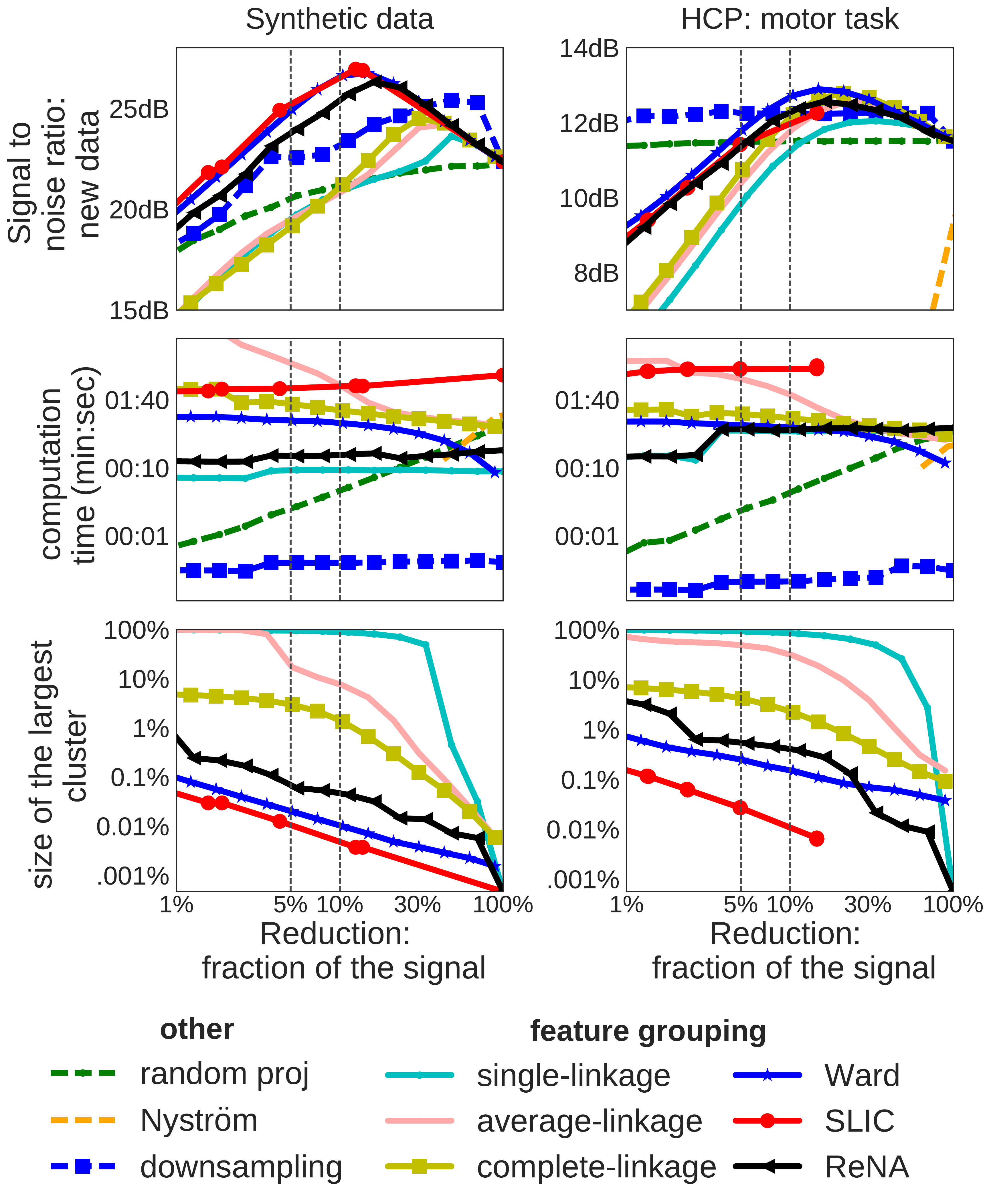} 
\caption{\textbf{Assessment of various approximation techniques on 
synthetic and brain imaging data:}
Evaluation of the performance varying the number $k$ of clusters.
\emph{(top)} Empirical measure of the distortion of the approximated 
distance.
For a fraction of the signal between $5\%$ and $30\%$ Ward, SLIC and ReNA 
present a denoising effect, improving the approximation of the distances.
In contrast, traditional agglomerative clustering fails to preserve
distances in the reduced space.
Downsampling displays an intermediate performance.
\emph{(center)} Regarding computation time, downsampling and random sampling 
outperform all the alternatives, followed by random projections and 
single-linkage.
The proposed method is almost as fast as single-linkage. 
\emph{(bottom)} Percolation behavior, measured via the size of the largest 
cluster. Ward, SLIC and ReNA are the best avoiding huge clusters. 
The vertical dashed lines indicate the useful value range for practical 
applications ($k \in \left[\lfloor p/20 \rfloor, \lfloor p/10\rfloor\right]$).
}
\label{fig:behavior_methods}
\end{figure}
\paragraph*{\textbf{Distortion}}
We want to test whether the reduction $\bPhi \ \bX$ of the noisy data is true 
to the uncorrupted signal $\bS$, 
\begin{equation}
\|\bPhi \ \bX_{*, i} - \bPhi \ \bX_{*, j}\|_2 \approx \|\bS_{*, i} - \bS_{*, 
j}\|_2, \ \forall (i, j)\in [n]^2.
\end{equation}
To do so, we split the uncorrupted signal matrix $\bS$ and the measured noisy 
data matrix $\bX$ into train and test sets, $(\bS^{\text{train}}, 
\bS^{\text{test}})$ and ($\bX^{\text{train}}, \bX^{\text{test}}$).
We learn a matrix $\bPhi$ on the noisy training data, $\bX^{\text{train}}$.
Then, we compare the pairwise distance between test samples of uncorrupted 
signal, $\bS^{\text{test}}$ to the pairwise distance between 
corresponding samples of noisy data after dimensionality reduction, 
$\bPhi\,\bX^{\text{test}}$. 
Finally, we report the relative distortion between these two distances
(for a full description see section 4, supplementary materials).
We carry out this experiment on two datasets: \emph{i)} synthetic data, and 
\emph{ii)} brain activation images (motor tasks) from the HCP dataset.

Fig.~\ref{fig:behavior_methods} (\emph{top}) presents the results on the
distortion.
Note that SLIC stops early in the range of the number of clusters.
In synthetic data, the clustering methods based on first-order linkage criteria 
(single, average, complete linkage) fail to represent the data accurately. 
By contrast, SLIC, Ward and ReNA achieve the best representation performance. 
These methods also show an expected denoising effect for
$\lfloor p/20 \rfloor < k < \lfloor p/10 
\rfloor$). In such range, the learned approximation matches approximately the 
smoothing kernel that characterizes the input signal. 
Downsampling also exhibits a denoising effect, needing more components than the 
non-percolating methods.
For the HCP dataset, the denoising effect is subtle, given that we do not
have access to noiseless signals.
Downsampling and random projections find a plateau in the relative distortion 
curve, meaning that the 
signal has a low entropy that is captured with only few components.
In both datasets, dimensionality reduction by random projections and
Nystr\"om fail to decrease the noise. This is because they guarantee 
good approximate distances, and hence represent also the noise.
\smallskip
\paragraph*{\textbf{Computation time}}
Fig.~\ref{fig:behavior_methods} (\emph{center}) gives
computation time for the different methods.
Dimension reduction by downsampling is the 
fastest, as it does not require any training and the 
computational time lies in the linear interpolation.
It is followed by the Nystr\"om approximation. 
While computation time of Nystr\"om approximation and random
projections increases with the reduction fraction of signal,
agglomerative clustering
approaches become faster, as they require less merges. Random projections
are faster than clustering approaches to reduce signals to a size
smaller than $30\%$ of their original size.
Among the clustering approaches, single-linkage and ReNA are the fastest, as
expected. 
Note that the cost of the clustering methods scales linearly with the number 
of samples, hence can be reduced by subsampling: using less data
to build the feature grouping.

\smallskip
\paragraph*{\textbf{Percolation behavior}}
As percolation is characterized by the occurrence of a huge cluster 
when $k$ decreases, we report the size of the largest cluster
varying the number $k$ of clusters on Fig. \ref{fig:behavior_methods} (\emph{bottom}).
Among the traditional agglomerative methods, single and average 
linkage display the worst behavior and quickly percolate.
Complete-linkage exhibits a more progressive behavior, with large
clusters that grow slowly in the small $k$ regime. 
On the other hand, Ward and SLIC are most resilient to percolation.
Indeed, they
are both known to create clusters of balanced size.
Finally, ReNA achieves a slightly worse performance, but mostly avoids huge 
clusters. 
\subsection{\label{sec:prediction}Use in prediction tasks}
To evaluate the denoising properties of dimension reduction, 
we now consider their use in prediction tasks.
We use linear estimators as they are standard in high dimensional 
problems. 
We consider $\ell_2$ and $\ell_1$ penalties.
For the $\ell_2$ case, the operator 
$\bPhif\transpose\bPhif$ acts like a kernel.
For such estimators, dimension reductions that preserve pairwise distance
are thus well theoretically motivated~\cite{rahimi2007}.

For each estimation problem, we use the relevant metric (explained 
variance\footnote{The explained variance is defined as $R^2 = 1 - 
\frac{\Var(\text{model} - \text{signal})}{\Var(\text{signal})}$} for regression 
and accuracy\footnote{Accuracy is defined by: $1 - 
\frac{\text{number of miss-classifications}}{\text{total number of samples}}$} 
for classification). 
We measure the performance of the pipeline: dimension reduction + estimator. 
Results are compared to those obtained without dimension reduction (raw 
data).
\subsubsection{Spatial approximation on a faces recognition task}
A classic pipeline to tackle face recognition
consists of first dimensionality reduction 
of the data followed by classifier training. 
Pipelines may include random projections, PCA, 
downsampling~\cite{wright2009}, or dictionary learning~\cite{xiang2011}.
Dimension reduction is motivated because varying illumination on a
subject with a fixed pose creates a low-dimensional subspace \cite{Basri2003}. 

We follow a study on reduced face representations~\cite{wright2009},
computing prediction accuracy for various feature-space dimensions 
$k \in \{30, 56, 120, 504\}$, corresponding to downsampling ratios of 
$\{1/32, 1/24, 1/16, 1/8\}$.
For the classifier, we use an $\ell_2$ or $\ell_1$ logistic 
regression with a multi class \emph{one-vs-rest} strategy and set the 
regularization parameter $\lambda$ by $10$-fold nested cross-validation. 
We measure prediction error with 50 iterations of cross-validation, 
randomly splitting half of each subject's images into train and test set.

\begin{figure}[bt]
\centering
\includegraphics[width=\linewidth]{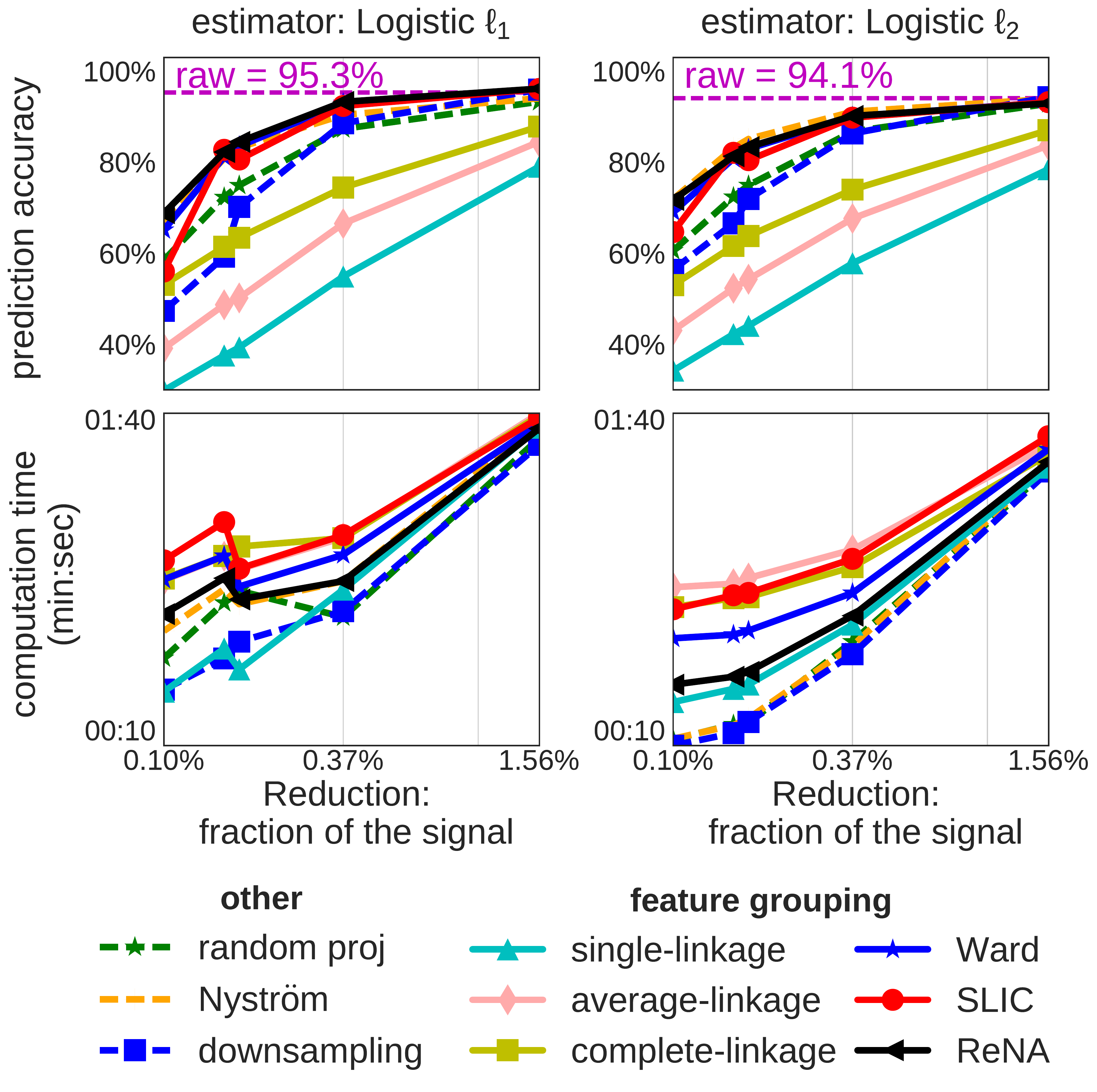}
\caption{\textbf{Face prediction accuracy for various approximation 
schemes:} Prediction accuracy as function of the feature space dimension 
obtained for various approximation schemes and classifiers for face
recognition on the extended Yale B dataset. 
The clustering methods finding balanced clusters need less features to have a 
fair performance, and they obtain significantly higher scores than the 
percolating methods.
}
\label{fig:faces_prediction}
\end{figure}

Fig.~\ref{fig:faces_prediction} reports the prediction accuracies.
For high reduction factors, Ward, Nystr\"om, and ReNA perform up to 
$10 \%$ better 
than random projections or downsampling: representations adjusted on the
data outperform data-independent reduction operators.
For raw data, without reduction,
prediction accuracy is around $95.3 \%$ and $94.1\%$ for the 
$\ell_1$ and $\ell_2$ penalization respectively. Similar performance is
obtained after reducing the signal by a factor of $64$
with random projections, Nystr\"om, downsampling, Ward, SLIC or ReNA. 
In contrast, single, average and complete linkage clustering fail to 
achieve the same performance.
This shows the importance of finding balanced clusters.

Regarding computation time, data reduction speeds up the convergence of
the logistic regression. Nystr\"om and downsampling 
are the fastest methods. 
Random projections, single-linkage, and ReNA follow,
all with similar performances. 
Average, complete linkage, Ward, and SLIC are slightly slower on this dataset.

\subsubsection{Trade-offs: prediction accuracy on a time budget}

Here, we examine the impact of the signal approximation on prediction
accuracy and prediction time. We use several datasets: in addition to
faces, anatomical and functional brain images. Supervised learning on
brain images~\cite{mitchell2004learning,schwartz2013} is a typical
setting that faces a rapid increase in dimensionality. Indeed, with
progresses in MRI, brain images are becoming bigger, leading to
computational bottlenecks. The Human Connectome Project (HCP) is
prototypical of these challenges, scanning 1\,200 subjects with
high-resolution protocols. We consider both anatomical brain images
(OASIS dataset, $n=403$ and $p=140\,000$) and functional brain images 
(HCP dataset, $n=8\,294$ and $p=250\,000$), with 3 different prediction
problems: age and gender prediction from anatomy, and discriminating 17
cognitive tasks from functional imaging.

We use the dimension reduction approaches to speed up predictor
training. We are interested in the total computation time needed to
learn a model: the cost of computing the compressed representation and
of training the classifier.
Based on prior experiments, we set $k = \lfloor p / 20
\rfloor$ for random projections, downsampling and clustering methods, and
$k = \lfloor n / 10 \rfloor$ for Nystr\"om, except for the faces dataset, 
where we use the $k = 504$ for all the methods. For classification, we
use a multinomial logistic regression with an $\ell_2$ penalty and an
l-BFGS solver and for regression we use a ridge.

\begin{figure}[bt]
\centering
\includegraphics[width=\linewidth]{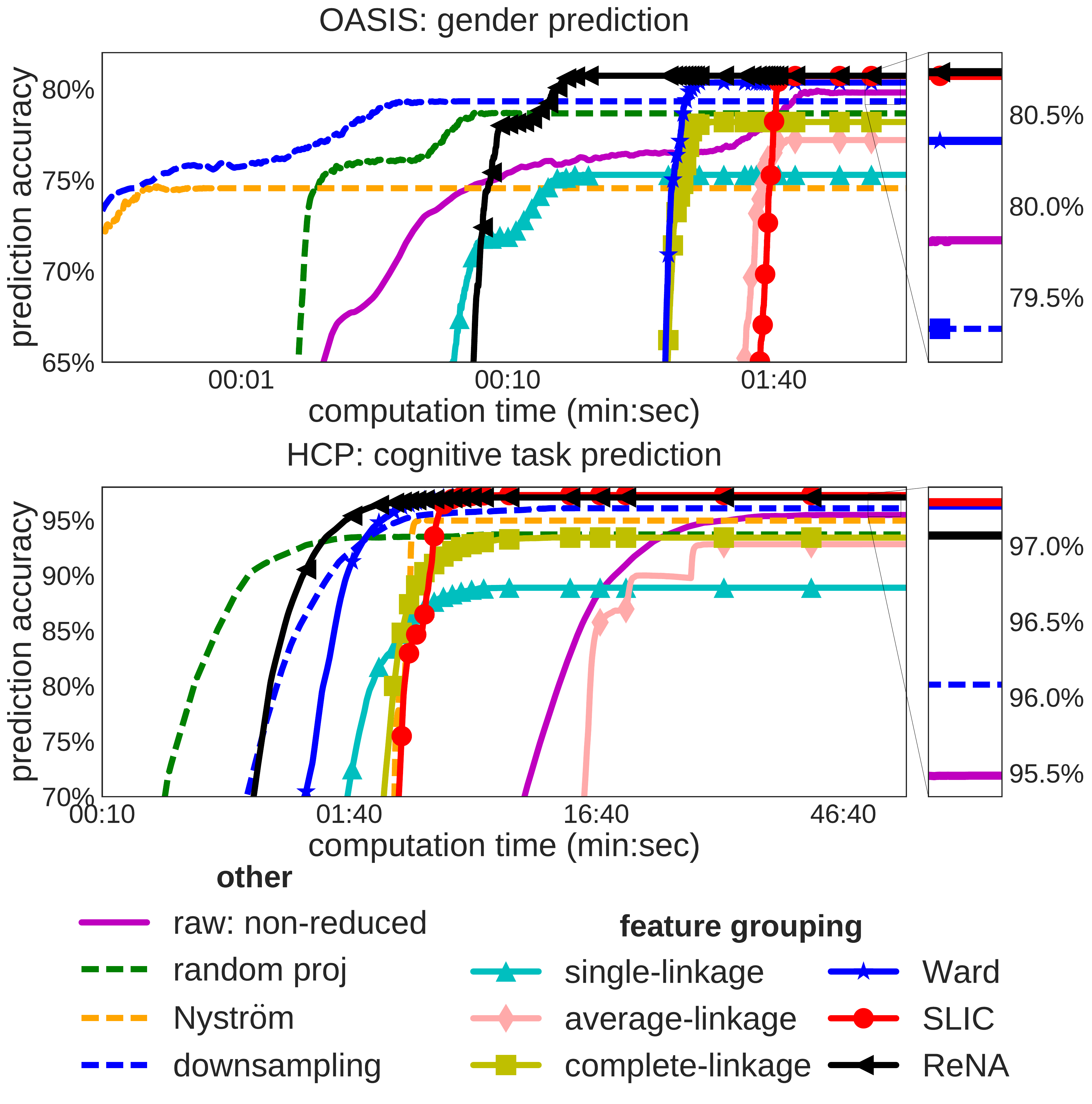}
\caption{\textbf{Computation time taken to reach a solution:} Quality of 
the 
fit of a $\ell_2$ penalized logistic regression as function of the computation 
time for a fixed number of clusters. 
In both datasets, Ward, SLIC and ReNA obtain significantly higher scores than 
estimation on non-reduced data with less computation time to reach a stable 
solution.
Note that the time displayed does include cluster computation.
}
\label{fig:time_to_predict}
\end{figure}

First, for a better understanding, we show on
Fig.~\ref{fig:time_to_predict} convergence of the logistic-regression
solver as a function of time on the two brain classification tasks. 
Time is spent in
learning the data reduction and iterations of the
solver. Interestingly, for some dimension reduction approaches, the
prediction reaches quickly a good accuracy regime, in particular for Ward
clustering and ReNA.
As with previous experiments, feature clustering with
single, average, and complete linkage lead to poor prediction. On the
opposite, SLIC, Ward, and ReNA give better prediction that non reduced
data, due to the denoising effect of feature clustering.

\begin{figure*}[bt]
\centering
\includegraphics[width=0.91 \linewidth]{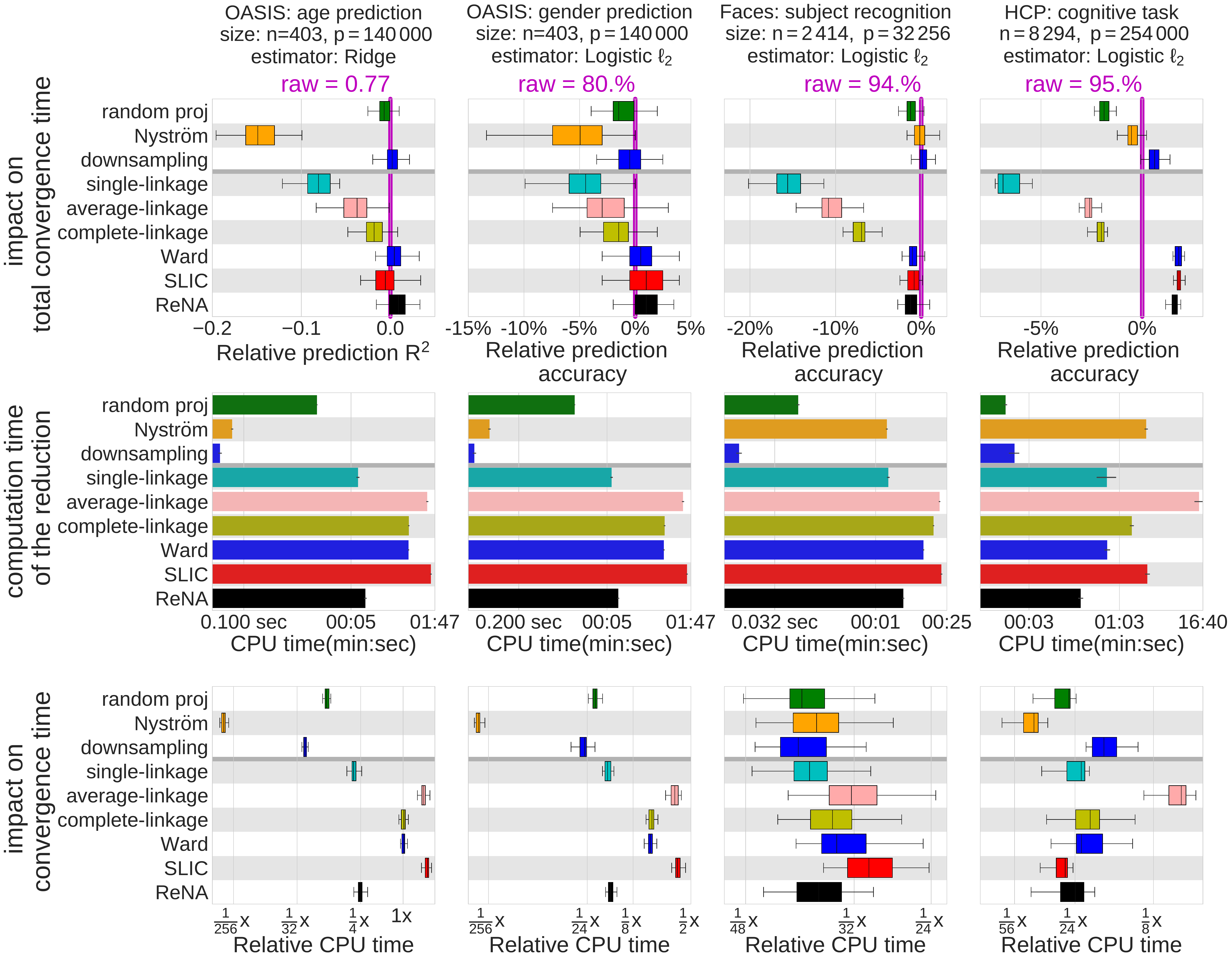} 
\caption{\textbf{Impact of reduction methods on prediction for various 
datasets:}
\emph{(Top)} Each bar represents the impact of the corresponding option on the 
prediction accuracy, relatively to the mean prediction with non-reduced data. 
Downsampling has the same performance as raw data. 
On the other hand, random projections, Nystr\"om, single, average and complete 
linkage algorithms are consistently the worst ones across datasets. 
Ward, SLIC and ReNA perform at least as good as non-reduced data.
\emph{(middle)} 
Regarding the computation time to find a reduction, single-linkage and ReNA are 
consistently the best among the clustering algorithms.   
Random projections perform better than Nystr\"om  when the number of samples is
large.
Downsampling is the fastest across datasets.
\emph{(Bottom)} 
The time to converge for single-linkage and ReNA is almost the same.
Average, complete-linkage and Ward are consistently the slowest.
SLIC performs well on large datasets.
Nystr\"om and random projections are the fastest across datasets.
Single-linkage and ReNA are the fastest clustering methods.
ReNA strikes a good trade off between time and prediction accuracy.
}
\label{fig:impact_on_prediction}
\end{figure*}

We then review systematically across datasets the impact
of the various data reductions on prediction accuracy, the time taken to
compute the reductions, and the total time to convergence
(Fig.~\ref{fig:impact_on_prediction}).
We find that dimension reduction with clustering algorithms that yield balanced
clusters (Ward, SLIC, and ReNA) achieves similar or better accuracy
than raw data while bringing drastic time savings.
Random projections and the percolating 
methods give consistently worse prediction accuracy than raw data.
On the OASIS dataset, downsampling, SLIC, and Ward achieve the same prediction 
accuracy as raw, and perform better than raw on other datasets.
Nystr\"om only performs as good as raw data on the faces dataset with an  
$\ell_2$ penalized logistic regression. ReNA has a slightly worse 
performance than raw only in this dataset, and displays a better performance 
than raw on the remaining datasets (p-value $< 10^{-4}$).
This illustrates the reduction of the spatial noise afforded by 
non-percolating clustering methods.

\begin{figure*}[bt]
\centering
\includegraphics[width=.9\linewidth]{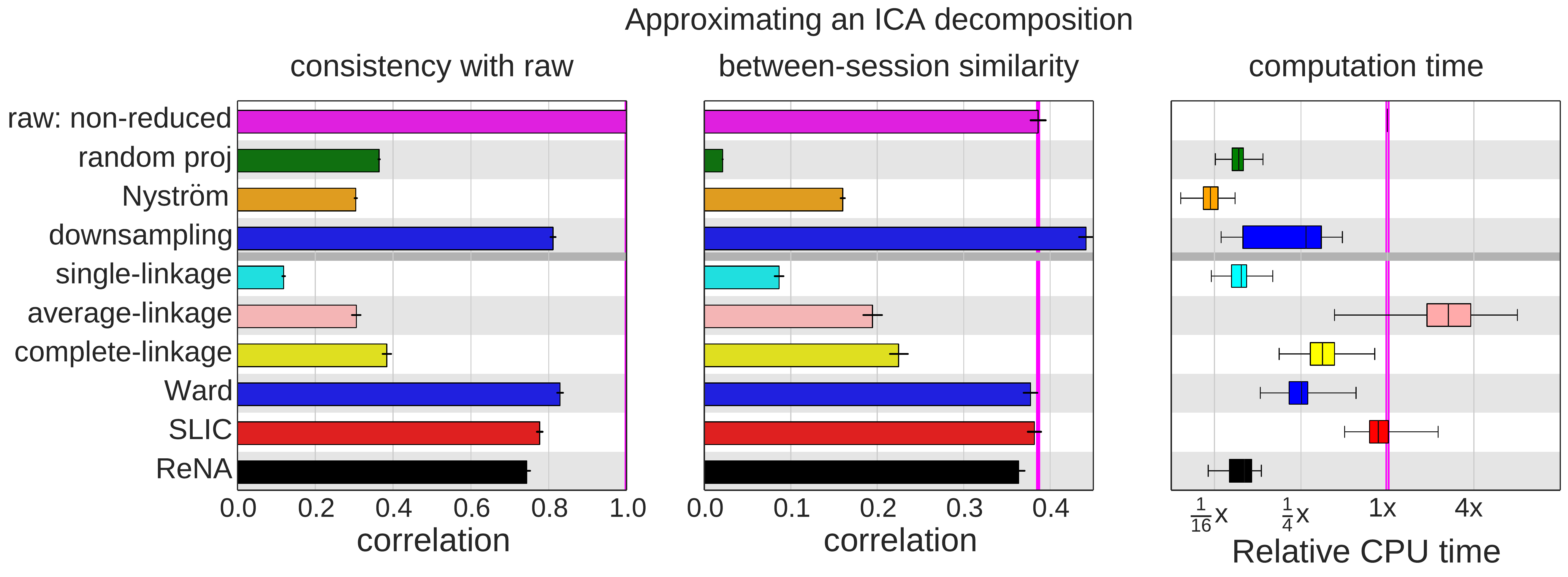}
\caption{\textbf{Spatial ICA reproducibility after dimension reduction:}
Reproducibility of $40$ spatial independent components on fMRI for
$93$ subjects,
with a fixed reduced dimension (see section 
\ref{sec:ICA}).
\emph{(Left)}
the similarity of downsampling, Ward, SLIC and ReNA with respect to the 
non-compressed components is high.
\emph{(Middle)} 
across two sessions, donwsampling yields components more consistent than 
raw data. 
Ward, SLIC and ReNA perform as well as raw data, while other approaches fail to 
do so.
\emph{(Right)}
regarding computational time, ReNA outperforms downsampling, Ward and SLIC, 
and performs as well as single-linkage and random projections. It is $16$
time faster than working on raw data.
}
\label{fig:ica_performance}
\end{figure*}

\subsection{\label{sec:ICA}Use in a spatial ICA task}
Aside from the $\ell_1$-penalized estimator, the data processing steps
studied above depend only on pairwise distances between samples. We
now investigate dimension reduction before an Independent Component
Analysis (ICA), which probes higher moments of the data distribution.
ICA is used routinely on resting-state fMRI to separate 
signal from noise or obtain functional networks~\cite{Smith2013}. 
We use $93$ subjects of the HCP data, with two rest fMRI sessions, each containing 
$1200$ brain images.

We compare ICA on the
raw data and after dimension reduction to 5\%
of the number of voxels ($k = \left\lfloor \frac{p}{20} \right\rfloor$).
For Nystr\"om, the dimension is set to 10\% of the number $n$ 
of samples ($k = \left\lfloor \frac{n}{10} \right\rfloor$).
In each subject, we extract $40$ independent components, a standard 
choice in the literature.
We investigate \emph{i)} how similar the components 
obtained are before and after reduction; \emph{ii)} how similar 
the components of session 1 and session 2 are with different reduction
approaches.
This second experiment gives a measure of the variability due to noise.
In both cases, we measure components similarity with
the absolute value of their correlation, and match them across sessions
with the Hungarian algorithm.

Fig.~\ref{fig:ica_performance} summarizes the use of dimension reduction 
in ICA of rest fMRI.
We find that the $40$ components are highly similar before and after 
data reduction with downsampling and Ward: the average absolute correlation 
greater than $0.8$. 
SLIC and ReNA have a slightly worse performance, with an average correlation 
greater than $0.74$.
On the other hand, single-linkage, average-linkage, complete-linkage,
Nyst\"om, 
and random projections do not recover the components  (average correlation $< 
0.4$). 
As expected, the components between sessions obtained by non-percolating 
clustering (Ward, SLIC, and ReNA) are similar to the original ones.  
Donwsampling improves the similarity with respect to raw: the estimation
problem is simpler and less noisy.
On the opposite, single, average, and complete linkage 
degrade the similarity: they lose signal due to the large cluster
created by percolation. 
Random projections and Nystr\"om perform poorly. 
Indeed, they average data across the images,
destroying the high-order moments of the data by creating
signals more Gaussian than the originals.
As a consequence, ICA cannot recover the 
sources derived from the original data. 
By contrast, the non-percolating clustering algorithms extract local
averages of the data, that preserve its
non-Gaussianity, as it has a spatial structure. 
Hence the spatial ICA is successful even though it has access to less samples.
Finally, dimensionality reduction using 
ReNA speeds up the total analysis by a factor of $15$.

\section{Summary and Discussion}

Fast dimension reduction is a crucial tool to tackle the rapid growth in 
datasets size, sample-wise and feature-wise.
In particular, grouping features is natural when
there is an underlying regularity in the observed signal, such as
spatial structure in images or a more general 
neighborhood structure connecting features.
We studied here a data-driven approach to perform feature grouping, where
groups are first learned from a fraction of the data using a
clustering algorithm, then used to build a compressed representation
for further analysis.

We showed that feature grouping can preserve well the pairwise
Euclidean distances between images.
This property makes it well suited for $\ell_2$-based 
algorithms, like shift-invariant kernel-based methods, or to approximate 
queries in information-retrieval settings.
We also clarified under which hypotheses this scheme leads to a
beneficial bias/variance compromise: as clustering adapts to the 
data, it reaches more optimal regimes than simple
downsampling-based compression.

Additionally, we proposed a linear-time graph-structured clustering
algorithm, ReNA, that is efficient with many clusters.
This algorithm iteratively performs 1-nearest neighbor grouping,
reduces the graph at each iteration, then averages the input features
and repeats the process until it reaches the desired number of
clusters.
We have shown empirically that it is fast, does not percolate, and
provides excellent performance in feature grouping for
dimension reduction of structured data.

Our experiments have shown that on moderate-to-large datasets,
non-percolating feature-grouping schemes (i.e. Ward, SLIC, and ReNA)
most often outperform state-of-the-art fast data-approximation
approaches for machine learning, namely random projection and random sampling.
Using these methods in a predictive pipeline increases the quality of
statistical estimations: they yield more accurate predictions than
with all features.
This indicates that feature grouping leads to a good approximation of
the data, capturing structure and reducing noise.
This denoising is due to the smoothness of the signal of interest: unlike the 
noise, the signal displays structure captured by feature grouping.

A key benefit of the ReNA clustering algorithm is that it is very
fast while avoiding percolation.
As a result, it gives impressive speed-ups for real-world multivariate 
statistical problems:  
often more than one order of magnitude.
Note that the computational cost of ReNA is linear in the number of 
samples, hence additional computation gains can be obtained by
sub-sampling its training data, as in Nystr\"om approaches.
In this work, we did not investigate the optimal choice of the number
$k$ of clusters, because we do not view compressed representations as a
meaningful model per se, but as an approximation to reduce data dimension 
without discarding too much information.
The range 
$k \in \left[ \left\lfloor \frac{p}{20} \right\rfloor, \left\lfloor 
\frac{p}{10} \right\rfloor \right]$
is a useful regime as it gives a good trade-off 
between computational efficiency and data fidelity. 
In our experiments, $k = \left\lfloor \frac{p}{20} \right\rfloor$
gave enough data fidelity for statistical analysis to perform at least as 
well as on raw data.
In this regime, Ward clustering gives slightly better approximations of
the original data, however it is slower, often by several orders of magnitude, hence
it is impractical.

We have shown that feature grouping is useful beyond
$\ell_2$-distance-based methods: it also gives good performance on
estimators relying on higher order moments (e.g. ICA) or sparsity 
($\ell_1$-based
regression or classification)\footnote{On the faces data for the $\ell_1$
estimator, ReNA performs as well as raw data.}.
As future work, it would be interesting to investigate the use of
ReNA-based feature grouping in expensive sparse algorithms, for instance
with sparse dictionary learning, where feature sub-sampling can give
large speed ups~\cite{mensch2016dictionary}.
Similarly, the combination of clustering, randomization, and sparsity 
has also been shown to be an effective regularization for some ill-posed
inverse problems~\cite{varoquaux2012icml,buhlmann2012}. 
This is all the more important that computation cost is a major 
roadblock to the adoption of such estimators.

An important aspect of feature grouping compared to other fast
dimension reductions, such as random projections, is that the features
of the reduced representation make sense for the application.
Consequently, the dimension reduction step can be inverted,
and any statistical analysis performed after reduction can be reported
with regard to the original signal.

Given that ReNA clustering is very fast, the proposed
featuring-grouping is an extremely promising avenue to speed up any
statistical analysis of large datasets where the information is in the
large-scale structure of the signal.
Such approach is crucial for domains where the resolution of the sensors
is rapidly increasing, in medical or biological imaging, genomics,
spectroscopy, or geospatial data.

\vskip 0.2in
\section*{Acknowledgments}
Data were provided in part by the Human Connectome Project, WU-Minn Consortium 
(Principal Investigators: D. Van Essen and K. Ugurbil; 1U54MH091657) 
funded by the 16 NIH Institutes and Centers that support the NIH Blueprint for 
Neuroscience Research; and by the McDonnell Center for Systems Neuroscience at 
Washington University.
Research leading to these results has received funding from the
European Union's Horizon 2020 Framework Programme for Research and
Innovation under Grant Agreement No 720270 (Human Brain Project SGA1).

\bibliographystyle{IEEEtran} 
\bibliography{biblio}

\begin{thebibliography}{10}
\providecommand{\url}[1]{#1}
\csname url@samestyle\endcsname
\providecommand{\newblock}{\relax}
\providecommand{\bibinfo}[2]{#2}
\providecommand{\BIBentrySTDinterwordspacing}{\spaceskip=0pt\relax}
\providecommand{\BIBentryALTinterwordstretchfactor}{4}
\providecommand{\BIBentryALTinterwordspacing}{\spaceskip=\fontdimen2\font plus
\BIBentryALTinterwordstretchfactor\fontdimen3\font minus
  \fontdimen4\font\relax}
\providecommand{\BIBforeignlanguage}[2]{{%
\expandafter\ifx\csname l@#1\endcsname\relax
\typeout{** WARNING: IEEEtran.bst: No hyphenation pattern has been}%
\typeout{** loaded for the language `#1'. Using the pattern for}%
\typeout{** the default language instead.}%
\else
\language=\csname l@#1\endcsname
\fi
#2}}
\providecommand{\BIBdecl}{\relax}
\BIBdecl

\bibitem{Amat2013}
F.~Amat, E.~W. Myers, and P.~J. Keller, ``Fast and robust optical flow for
  time-lapse microscopy using super-voxels,'' \emph{Bioinformatics}, vol.~29,
  pp. 373--80, 2013.

\bibitem{VanEssen2012}
D.~V. Essen, K.~Ugurbil, E.~Auerbach, D.~Barch, T.~Behrens, R.~Bucholz,
  A.~Chang, L.~Chen, M.~Corbetta, S.~Curtiss, S.~D. Penna, D.~Feinberg,
  M.~Glasser, N.~Harel, A.~Heath, L.~Larson-Prior, D.~Marcus, G.~Michalareas,
  S.~Moeller, R.~Oostenveld, S.~Petersen, F.~Prior, B.~Schlaggar, S.~Smith,
  A.~Snyder, J.~Xu, and E.~Yacoub, ``The human connectome project: A data
  acquisition perspective,'' \emph{NeuroImage}, vol.~62, pp. 2222--2231, 2012.

\bibitem{Overbeek2000}
R.~Overbeek, N.~Larsen, G.~D. Pusch, M.~D'Souza, E.~S. Jr, N.~Kyrpides,
  M.~Fonstein, N.~Maltsev, and E.~Selkov, ``{WIT}: integrated system for
  high-throughput genome sequence analysis and metabolic reconstruction,''
  \emph{Nucleic Acids Research}, vol.~28, pp. 123--125, 2000.

\bibitem{Cole2013}
J.~R. Cole, Q.~Wang, J.~A. Fish, B.~Chai, D.~M. McGarrell, Y.~Sun, C.~T. Brown,
  A.~Porras-Alfaro, C.~R. Kuske, and J.~M. Tiedje, ``Ribosomal database
  project: data and tools for high throughput rrna analysis,'' \emph{Nucleic
  Acids Research}, 2013.

\bibitem{addair2014seismic}
T.~Addair, D.~Dodge, W.~Walter, and S.~Ruppert, ``Large-scale seismic signal
  analysis with hadoop,'' \emph{Computers \& Geosciences}, vol.~66, p. 145,
  2014.

\bibitem{jegou2010improving}
H.~J{\'e}gou, M.~Douze, and C.~Schmid, ``Improving bag-of-features for large
  scale image search,'' \emph{Int. J. Comput. Vision}, vol.~87, pp. 316--336,
  2010.

\bibitem{hoyos2015}
A.~Hoyos-Idrobo, Y.~Schwartz, G.~Varoquaux, and B.~Thirion, ``Improving sparse
  recovery on structured images with bagged clustering,'' \emph{IEEE PRNI},
  2015.

\bibitem{Chakrabarti:2002:LAD:568518.568520}
K.~Chakrabarti, E.~Keogh, S.~Mehrotra, and M.~Pazzani, ``Locally adaptive
  dimensionality reduction for indexing large time series databases,''
  \emph{ACM Trans. Database Syst.}, vol.~27, pp. 188--228, 2002.

\bibitem{keogh2001dimensionality}
E.~Keogh, K.~Chakrabarti, M.~Pazzani, and S.~Mehrotra, ``Dimensionality
  reduction for fast similarity search in large time series databases,''
  \emph{Knowledge and information Systems}, vol.~3, p. 263, 2001.

\bibitem{rahimi2007}
A.~Rahimi and B.~Recht, ``Random features for large-scale kernel machines,'' in
  \emph{NIPS}, 2007.

\bibitem{Ailon2009}
N.~Ailon and B.~Chazelle, ``The fast {J}ohnson-{L}indenstrauss transform and
  approximate nearest neighbors,'' \emph{SIAM J. Comput.}, vol.~39, pp.
  302--322, 2009.

\bibitem{Halko2011}
N.~Halko, P.~G. Martinsson, and J.~A. Tropp, ``Finding structure with
  randomness: Probabilistic algorithms for constructing approximate matrix
  decompositions,'' \emph{SIAM Rev.}, vol.~53, pp. 217--288, 2011.

\bibitem{GittensM13}
A.~Gittens and M.~W. Mahoney, ``Revisiting the {N}ystr\"om method for improved
  large-scale machine learning.'' in \emph{ICML}, 2013, pp. 567--575.

\bibitem{mahoney2009matrix}
M.~W. Mahoney and P.~Drineas, ``{CUR} matrix decompositions for improved data
  analysis,'' \emph{PNAS}, vol. 106, pp. 697--702, 2009.

\bibitem{Johnson1984}
W.~Johnson and J.~Lindenstrauss, ``Extensions of {L}ipschitz mappings into a
  {H}ilbert space,'' in \emph{Conference in modern analysis and probability},
  ser. Contemporary Mathematics.\hskip 1em plus 0.5em minus 0.4em\relax
  American Mathematical Society, 1984, vol.~26, pp. 189--206.

\bibitem{lin2007}
J.~Lin, E.~Keogh, L.~Wei, and S.~Lonardi, ``Experiencing {SAX}: a novel
  symbolic representation of time series,'' \emph{Data Mining and knowledge
  discovery}, pp. 107--144, 2007.

\bibitem{shuman2013emerging}
D.~I. Shuman, S.~K. Narang, P.~Frossard, A.~Ortega, and P.~Vandergheynst, ``The
  emerging field of signal processing on graphs: Extending high-dimensional
  data analysis to networks and other irregular domains,'' \emph{Signal
  Processing Magazine, IEEE}, vol.~30, pp. 83--98, 2013.

\bibitem{stauffer1992}
D.~Stauffer and A.~Aharony, \emph{{Introduction to Percolation Theory}}.\hskip
  1em plus 0.5em minus 0.4em\relax Oxford University Press, New York, 1971.

\bibitem{hedge2015}
C.~Hedge, A.~C. Sankaranarayanan, W.~Yin, and R.~G. Baraniuk, ``{N}u{M}ax: a
  convex approach for learning near-isometric linear embeddings,'' \emph{IEEE
  Trans. Signal Process.}, vol.~63, p. 6109, 2015.

\bibitem{LuDFU13}
Y.~Lu, P.~S. Dhillon, D.~P. Foster, and L.~H. Ungar, ``Faster ridge regression
  via the subsampled randomized hadamard transform.'' in \emph{NIPS}, 2013, pp.
  369--377.

\bibitem{achlioptas2003}
D.~Achlioptas, ``Database-friendly random projections:
  {J}ohnson-{L}indenstrauss with binary coins,'' \emph{Journal of Computer and
  System Sciences}, vol.~66, pp. 671--687, 2003.

\bibitem{tropp2011}
J.~A. Tropp, ``Improved analysis of the subsampled randomized hadamard
  transform,'' \emph{Advances in Adaptive Data Analysis}, vol.~3, pp. 115--126,
  2011.

\bibitem{Williams01usingthe}
C.~Williams and M.~Seeger, ``Using the {N}ystr\"om method to speed up kernel
  machines,'' in \emph{NIPS}, 2001, pp. 682--688.

\bibitem{Rudi2015}
A.~Rudi, R.~Camoriano, and L.~Rosasco, ``Less is more: {N}ystr\"om
  computational regularization,'' in \emph{NIPS}, 2015, pp. 1648--1656.

\bibitem{buhlmann2012}
P.~{B{\"u}hlmann}, P.~{R{\"u}timann}, S.~{van de Geer}, and C.-H. {Zhang},
  ``{Correlated variables in regression: clustering and sparse estimation},''
  \emph{Journal of Statistical Planning and Inference}, vol. 143, p. 1835,
  2013.

\bibitem{le2015flexible}
L.~Le~Magoarou and R.~Gribonval, ``Flexible multilayer sparse approximations of
  matrices and applications,'' \emph{IEEE Journal of Selected Topics in Signal
  Processing}, vol.~10, pp. 688--700, 2016.

\bibitem{Achanta2012}
R.~Achanta, A.~Shaji, K.~Smith, A.~Lucchi, P.~Fua, and S.~S\"usstrunk, ``{SLIC}
  superpixels compared to state-of-the-art superpixel methods,'' \emph{IEEE
  Trans. Pattern Anal. Mach. Intell.}, vol.~34, p. 2274, 2012.

\bibitem{arthur2006kmeans}
D.~Arthur and S.~Vassilvitskii, ``How slow is the k-means method?'' in
  \emph{Annual Symposium on Computational Geometry}, 2006, p. 144.

\bibitem{hastie09statisticallearning}
T.~Hastie, R.~Tibshirani, and J.~Friedman, \emph{The Elements of Statistical
  Learning}, 2nd~ed., ser. Springer Series in Statistics.\hskip 1em plus 0.5em
  minus 0.4em\relax New York, NY, USA: Springer New York Inc., 2009.

\bibitem{ward1963}
J.~H. Ward, ``Hierarchical grouping to optimize an objective function,''
  \emph{J. Am. Stat. Ass.}, vol.~58, p. 236, 1963.

\bibitem{Mullner2011}
D.~M\"ullner, ``Modern hierarchical, agglomerative clustering algorithms,''
  \emph{ArXiv e-prints}, 2011.

\bibitem{mirkin1998mathematical}
B.~Mirkin, ``Mathematical classification and clustering: From how to what and
  why,'' in \emph{Classification, data analysis, and data highways}.\hskip 1em
  plus 0.5em minus 0.4em\relax Springer, 1998, pp. 172--181.

\bibitem{rosenfeld1966sequential}
A.~Rosenfeld and J.~L. Pfaltz, ``Sequential operations in digital picture
  processing,'' \emph{Journal of the ACM (JACM)}, vol.~13, p. 471, 1966.

\bibitem{Penrose1995}
M.~Penrose, ``Single linkage clustering and continuum percolation,''
  \emph{Journal of Multivariate Analysis}, vol.~53, pp. 94 -- 109, 1995.

\bibitem{thirion2014}
B.~Thirion, G.~Varoquaux, E.~Dohmatob, and J.-B. Poline, ``Which f{MRI}
  clustering gives good brain parcellations?'' \emph{Frontiers in
  Neuroscience}, vol.~8, 2014.

\bibitem{Felzenszwalb2004}
P.~F. Felzenszwalb and D.~P. Huttenlocher, ``Efficient graph-based image
  segmentation,'' \emph{Int. J. Comput. Vision}, vol.~59, p. 167, 2004.

\bibitem{Eppstein1997}
D.~Eppstein, M.~S. Paterson, and F.~Yao, ``On nearest-neighbor graphs".
  discrete and computational geometry,'' vol.~17, pp. 263--282, 1997.

\bibitem{Ester96adensity-based}
M.~Ester, H.~peter Kriegel, J.~Sander, and X.~Xu, ``A density-based algorithm
  for discovering clusters in large spatial databases with noise.''\hskip 1em
  plus 0.5em minus 0.4em\relax AAAI Press, 1996, pp. 226--231.

\bibitem{MAIER20091749}
M.~Maier, M.~Hein, and U.~von Luxburg, ``Optimal construction of
  k-nearest-neighbor graphs for identifying noisy clusters,'' \emph{Theoretical
  Computer Science}, vol. 410, no.~19, pp. 1749 -- 1764, 2009, algorithmic
  Learning Theory.

\bibitem{Teng2007}
S.-H. Teng and F.~F. Yao, ``k-nearest-neighbor clustering and percolation
  theory,'' \emph{Algorithmica}, vol.~49, pp. 192--211, 2007.

\bibitem{Pearce05animproved}
D.~J. Pearce, ``An improved algorithm for finding the strongly connected
  components of a directed graph,'' Tech. Rep., 2005.

\bibitem{tarjan1972depth}
R.~Tarjan, ``Depth-first search and linear graph algorithms,'' \emph{SIAM
  journal on computing}, vol.~1, pp. 146--160, 1972.

\bibitem{wright2009}
J.~Wright, A.~Yang, A.~Ganesh, S.~Sastry, and Y.~Ma, ``Robust face recognition
  via sparse representation,'' \emph{IEEE Trans. Pattern Anal. Mach. Intell.},
  vol.~31, pp. 210--227, 2009.

\bibitem{marcus2007}
D.~S. Marcus, T.~H. Wang, J.~Parker, J.~G. Csernansky, J.~C. Morris, and R.~L.
  Buckner, ``Open access series of imaging studies ({OASIS}): cross-sectional
  {MRI} data in young, middle aged, nondemented, and demented older adults.''
  \emph{J Cogn Neurosci}, vol.~19, pp. 1498--1507, 2007.

\bibitem{Chau2005408}
W.~Chau and A.~R. McIntosh, ``The talairach coordinate of a point in the {MNI}
  space: how to interpret it,'' \emph{NeuroImage}, vol.~25, pp. 408--416, 2005.

\bibitem{barch2013}
D.~M. Barch, G.~C. Burgess, M.~P. Harms, S.~E. Petersen, B.~L. Schlaggar,
  M.~Corbetta, M.~F. Glasser, S.~Curtiss, S.~Dixit, C.~Feldt, D.~Nolan,
  E.~Bryant, T.~Hartley, O.~Footer, J.~M. Bjork, R.~Poldrack, S.~Smith,
  H.~Johansen-Berg, A.~Z. Snyder, D.~C.~V. Essen, and W.~U.-M.~H. Consortium,
  ``Function in the human connectome: task-f{MRI} and individual differences in
  behavior.'' \emph{Neuroimage}, vol.~80, pp. 169--189, 2013.

\bibitem{glass13}
M.~F. Glasser, S.~N. Sotiropoulos, J.~A. Wilson, T.~S. Coalson, B.~Fischl,
  J.~L. Andersson, J.~Xu, S.~Jbabdi, M.~Webster, and J.~R. Polimeni, ``The
  minimal preprocessing pipelines for the human connectome project,''
  \emph{Neuroimage}, vol.~80, pp. 105--124, 2013.

\bibitem{Pedregosa2011}
F.~Pedregosa, G.~Varoquaux, A.~Gramfort, V.~Michel, B.~Thirion, O.~Grisel,
  M.~Blondel, P.~Prettenhofer, R.~Weiss, V.~Dubourg, J.~Vanderplas, A.~Passos,
  D.~Cournapeau, M.~Brucher, M.~Perrot, and E.~Duchesnay, ``Scikit-learn:
  Machine learning in {Python},'' \emph{Journal of Machine Learning Research},
  vol.~12, p. 2825, 2011.

\bibitem{scikit-image}
S.~van~der Walt, J.~L. {S}ch\"onberger, J.~{Nunez-Iglesias}, F.~{B}oulogne,
  J.~D. {W}arner, N.~{Y}ager, E.~{G}ouillart, T.~{Y}u, and the scikit-image
  contributors, ``Scikit-image: image processing in {P}ython,'' \emph{PeerJ},
  vol.~2, p. e453, 2014.

\bibitem{jones2014scipy}
E.~Jones, T.~Oliphant, and P.~Peterson, ``{SciPy}: open source scientific tools
  for {Python},'' 2014.

\bibitem{xiang2011}
Z.~J. Xiang, H.~Xu, and P.~J. Ramadge, ``Learning sparse representations of
  high dimensional data on large scale dictionaries,'' in \emph{NIPS}, 2011,
  pp. 900--908.

\bibitem{Basri2003}
R.~Basri and D.~Jacobs, ``Lambertian reflection and linear subspaces,''
  \emph{IEEE Trans. Pattern Anal. Mach. Intell.}, vol.~25, p. 218, 2003.

\bibitem{mitchell2004learning}
T.~M. Mitchell, R.~Hutchinson, R.~S. Niculescu, F.~Pereira, X.~Wang, M.~Just,
  and S.~Newman, ``Learning to decode cognitive states from brain images,''
  \emph{Machine Learning}, vol.~57, p. 145, 2004.

\bibitem{schwartz2013}
Y.~Schwartz, B.~Thirion, and G.~Varoquaux, ``Mapping cognitive ontologies to
  and from the brain,'' in \emph{NIPS}, 2013.

\bibitem{Smith2013}
S.~M. Smith, C.~F. Beckmann, J.~Andersson, E.~J. Auerbach, J.~Bijsterbosch,
  G.~Douaud, E.~Duff, D.~A. Feinberg, L.~Griffanti, M.~P. Harms, M.~Kelly,
  T.~Laumann, K.~L. Miller, S.~Moeller, S.~Petersen, J.~Power,
  G.~Salimi-Khorshidi, A.~Z. Snyder, A.~T. Vu, M.~W. Woolrich, J.~Xu,
  E.~Yacoub, K.~U\u{g}urbil, D.~C.~V. Essen, and M.~F. Glasser, ``Resting-state
  f{MRI} in the human connectome project,'' \emph{NeuroImage}, vol.~80, pp.
  144--168, 2013.

\bibitem{mensch2016dictionary}
A.~Mensch, J.~Mairal, B.~Thirion, and G.~Varoquaux, ``Dictionary learning for
  massive matrix factorization,'' \emph{ICML}, 2016.

\bibitem{varoquaux2012icml}
G.~Varoquaux, A.~Gramfort, and B.~Thirion, ``Small-sample brain mapping: sparse
  recovery on spatially correlated designs with randomization and clustering,''
  in \emph{ICML}, 2012, p. 1375.

\end{thebibliography}


\begin{thebibliography}{1}
\providecommand{\url}[1]{#1}
\csname url@samestyle\endcsname
\providecommand{\newblock}{\relax}
\providecommand{\bibinfo}[2]{#2}
\providecommand{\BIBentrySTDinterwordspacing}{\spaceskip=0pt\relax}
\providecommand{\BIBentryALTinterwordstretchfactor}{4}
\providecommand{\BIBentryALTinterwordspacing}{\spaceskip=\fontdimen2\font plus
\BIBentryALTinterwordstretchfactor\fontdimen3\font minus
  \fontdimen4\font\relax}
\providecommand{\BIBforeignlanguage}[2]{{%
\expandafter\ifx\csname l@#1\endcsname\relax
\typeout{** WARNING: IEEEtran.bst: No hyphenation pattern has been}%
\typeout{** loaded for the language `#1'. Using the pattern for}%
\typeout{** the default language instead.}%
\else
\language=\csname l@#1\endcsname
\fi
#2}}
\providecommand{\BIBdecl}{\relax}
\BIBdecl

\bibitem{Mahoney:2011:RAM:2185807.2185808}
M.~W. Mahoney, ``Randomized algorithms for matrices and data,'' \emph{Found.
  Trends Mach. Learn.}, vol.~3, pp. 123--224, 2011.

\bibitem{Katzgraber2010}
H.~G. Katzgraber, ``Random numbers in scientific computing: An introduction,''
  \emph{CoRR}, vol. abs/1005.4117, 2010.

\end{thebibliography}
\vspace{-1.0cm}
\begin{IEEEbiography}
[{\includegraphics[width=1in,height=1.25in,clip,keepaspectratio]{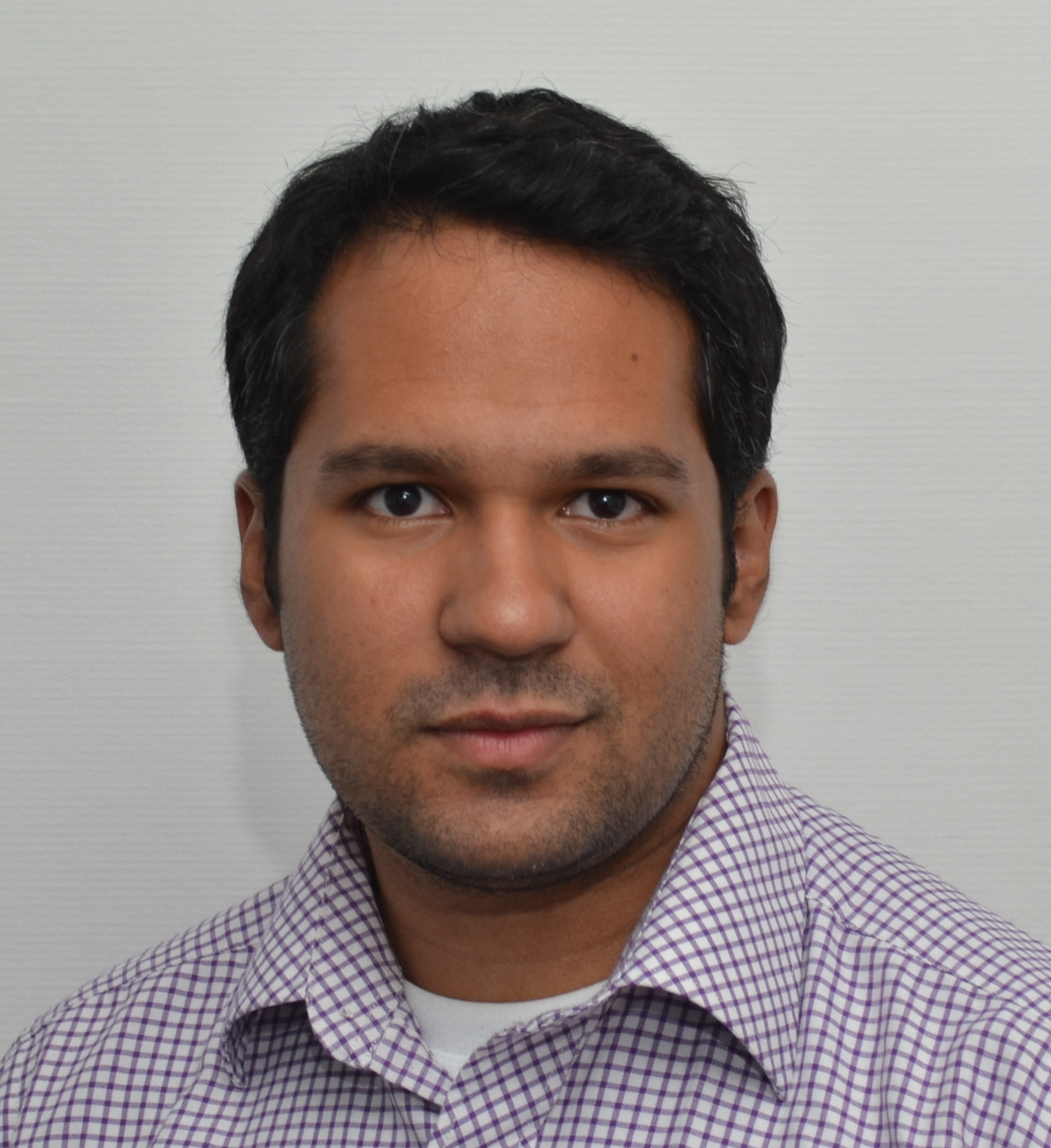}}]
{Andr\'es Hoyos Idrobo} received his is B.S degree in Electronic and Biomedical 
engineering from the University Aut\'onoma de occidente, Colombia, and
his M.S.c degree in applied mathematics from \'Ecole Normale 
Sup\'erieure de Cachan, France. 
He has a Ph.D computer science from University Paris-Saclay.
His research focus on faster methods for brain activity decoding,
in particular, randomized algorithms.
\end{IEEEbiography}
\vspace{-0.8cm}
\begin{IEEEbiography}%
[{\includegraphics[width=1in,height=1in,clip,keepaspectratio]{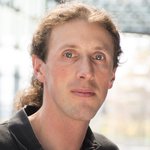}}]%
{Ga\"el Varoquaux} is a computer-science researcher at INRIA. 
He develops statistical learning for functional neuroimaging, 
applied to cognitive brain mapping and brain pathologies. 
In addition, he is invested in data-science software, as project lead for 
scikit-learn, joblib, and 
nilearn. 
He has a PhD in quantum physics and is a graduate from \'Ecole Normale 
Sup\'erieure, Paris.%
\end{IEEEbiography}
\vspace{-1.0cm}
\begin{IEEEbiography}%
[{\includegraphics[width=1in,height=1.25in,clip,keepaspectratio]{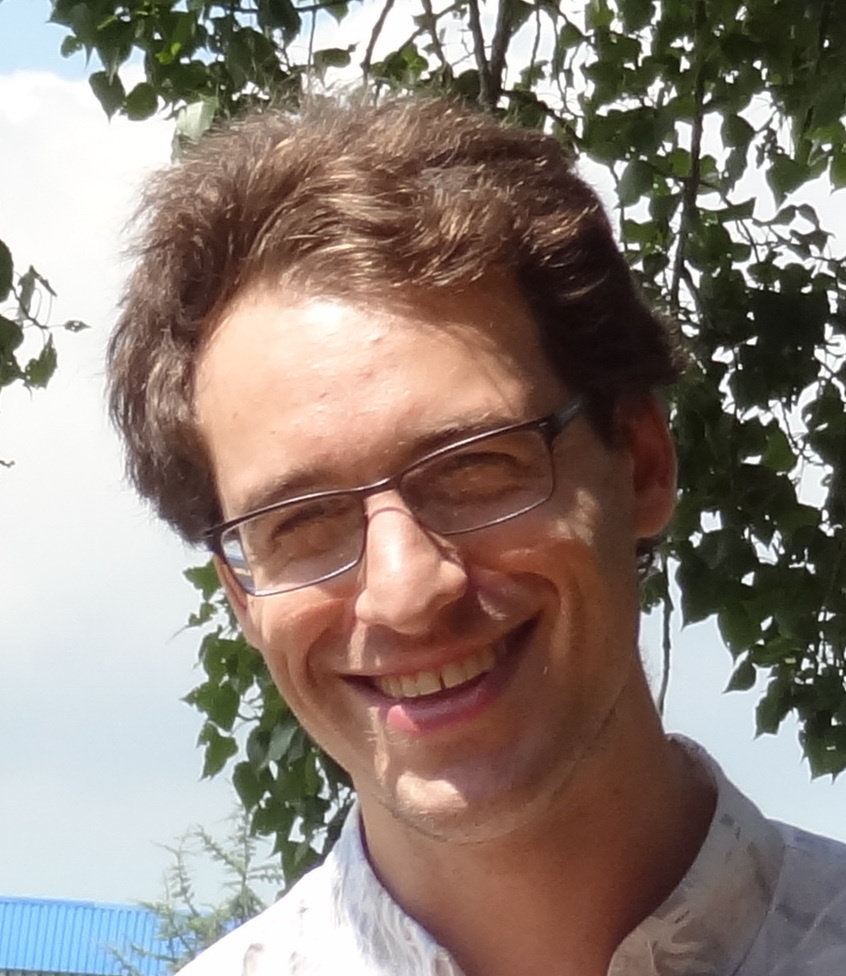}}]%
{Jonas KAHN} is a mathematician with interests ranging from fundamental
subjects such as operator algebras or random geometric spaces to
statistics and applications. On the latter side, he has worked in
particular on quantum statistics, and compressed sensing with physical
constraints. He has two PhD in mathematics, from Orsay and Leiden, and
has graduated from \'Ecole Normale Sup\'erieure.
\end{IEEEbiography}
\vspace{-0.8cm}
\begin{IEEEbiography}
[{\includegraphics[width=1in,height=1.25in,clip,keepaspectratio]{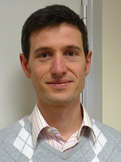}}]
{Bertrand Thirion} is the principal investigator of the Parietal team
(Inria-CEA) within the main French Neuroimaging center, Neurospin.  He
is deputy scientific leader of the Inria Saclay research center.  His
main research interests address the design of machine learning and
statistical analysis techniques for neuroimaging, with
applications related to the study of human vision, brain disease
diagnosis, population imaging and cognitive brain mapping. 

\end{IEEEbiography}
\vfill

\end{document}


\title{Supplementary materials: Recursive nearest agglomeration (ReNA) -- fast 
clustering for approximation of structured signals}

\maketitle 

\section{\label{sec:nystrom}Nystr\"om feature mapping}
Here, we present the standard implementation of the Nystr\"om approximation for 
linear kernels.
Algorithm \ref{al:Nystrom} allows to build a data-driven feature mapping
that is used to reduce the dimensionality of the data matrix.
The algorithm is summarized as follows: first, we select images uniformly at 
random\footnote{We can use other sampling probabilities (e.g. the leverage 
score \cite{Mahoney:2011:RAM:2185807.2185808}), but by sampling uniformly we 
are assuming a regular structure}, then we calculate the kernel of these 
samples and use it to normalize the selected images.
\begin{algorithm}[h!]{}
\caption{Nystr\"om: Learning the feature mapping}
\label{al:Nystrom}
\begin{algorithmic}[1]
\REQUIRE The training data matrix $\bX \in \reals^{p\times n}$, number $k$ of 
components, where $k < n$.
\ENSURE The feature mapping $\bPhin \in \reals^{k\times p}$\\ 
\STATE $\br \leftarrow$ Generate uniform sampling of $k$ components\\
\STATE $\bX_{*, \br} \in \reals^{p\times k}$ \COMMENT{Subsample of $k$ 
columns}\\
\STATE $\tilde{\bK} = \bXT_{*, \br} \bX_{*, \br}$ \COMMENT{Kernel 
matrix of the subsampled data}\\
\STATE $\bPhin = \tilde{\bK}^{-1/2} \bX_{*, \br}^T$ \COMMENT{Normalization: 
via SVD}\\
\RETURN $\bPhin$
\end{algorithmic}
\end{algorithm}

\section{Proof of Lemma 2.2} 

\label{sec:inertia}

In this appendix, we present the arithmetic manipulation necessary to prove the 
first part of the Lemma 2.2.
As $\bPhif\transpose \bPhif$ is an orthogonal operator, Eq. 6
corresponds to an orthogonal decomposition.
But, for the sake of clarity we include it in this appendix.
\begin{corollary}
Let $\bx \in \reals^p$ be a signal, and $\bPhif$ be a feature-grouping 
matrix, the following holds  
\begin{mequation}
\|\bx\|_2^2 - \sum_{q=1}^{k}\left\|\bx_{\cC_q} - \frac{(\bPhif \ 
\bx)_q}{\sqrt{|\cC_q|}}\right\|_2^2 = \left\|\bPhif \ \bx\right\|_2^2  \leq 
\|\bx\|_2^2.
\label{eq:coro_distortion}
\end{mequation}
\label{coro:distortion}
\end{corollary}

\begin{proof}
We start by writing down the $\ell_2$ norm of the data vector $\bx$ for every 
point inside all the clusters $\{\cC_q\}_{q=1}^k$. 
Then, we perform simple manipulations, as follows
\begin{mequation}
\begin{split}
\|\bx\|_2^2 & = \sum_{q=1}^{k} \sum_{i \in \cC_q} \bx_i^2 \\ 
& = \sum_{q=1}^{k} \sum_{i \in \cC_q} \left(\bx_i^2 + 
\frac{(\bPhif \ \bx)_q^2}{|\cC_q|} - \frac{(\bPhif \ \bx)_q^2}{|\cC_q|}\right)\\
& = \sum_{q=1}^{k} (\bPhif \ \bx)_q^2 + \sum_{q=1}^{k} \sum_{i \in \cC_q} 
\left(\bx_i^2 - \frac{(\bPhif \ \bx)_q^2}{|\cC_q|}\right) \\
& = \left\|\bPhif \ \bx\right\|_2^2 + \sum_{q=1}^{k}\left\|\bx_{\cC_q} - 
\frac{(\bPhif \ \bx)_q}{\sqrt{|\cC_q|}}\right\|_2^2.
\end{split}
\label{eq:distortion_appendix}
\end{mequation}

Finally, the right hand inequality of Eq.\ref{eq:coro_distortion}
comes naturally after scaling each cluster (the non-zero singular values are 
set to 
$1$). 
\hfill$\square$
\end{proof}
\begin{proof}
The corollary \ref{coro:distortion} shows that the lower bound of the 
representation is only affected by the inertia (see left hand side of 
Eq.\ref{eq:coro_distortion}). 
So, the worst case corresponds to the upper bound of the inertia 
$M(\bx)=\sum_{q=1}^{k}m_q(\bx)$. 
Then, the inertia of each cluster can be bounded as follows 
\begin{mequation}
\label{eq:first_bound_inertia}
\begin{split}
m_q(\bx)  & = \left\|\bx_{\cC_q} - \frac{(\bPhif \ 
\bx)_q}{\sqrt{|\cC_q|}}\right\|_2^2 \\
& = \sum_{j\in \cC_q} \left(\bx_j - \frac{1}{|\cC_q|}\sum_{i\in \cC_q}  
\bx_i\right)^2\\
& \leq |\cC_q| \max_{j\in \cC_q} \left|\bx_j - \frac{1}{|\cC_q|}\sum_{i\in 
\cC_q}  \bx_i\right|^2\\
\end{split}
\end{mequation}
The last inequality corresponds to the worst case for the sum inside the 
cluster.

Let constrain our analysis to L-smooth signals $\bx \in \reals^p$ structured by 
graph $\cG$ (see Definition 2.1).  
Under this assumption, we can bound the inertia of each cluster as follows:
\begin{mequation}
\begin{split}
m_q(\bx) &\leq |\cC_q| \max_{j\in \cC_q} \left|\bx_j - 
\frac{1}{|\cC_q|}\sum_{i\in \cC_q}  \bx_i\right|^2 \\
&\leq |\cC_q|\,L^2\,\sup_{i,j \in \cC_q}\text{dist}_\cG(v_i, v_j)^2\\
&=  L^2\,|\cC_q|\,\diamG(\cC_q)^2,
\end{split}
\label{eq:bound_inertia_cluster}
\end{mequation}
where the second inequality follows from the pairwise L-Lipschitz condition.
Finally, plugging Eq.\ref{eq:bound_inertia_cluster} into 
Eq.\ref{eq:distortion_appendix} we have:
\begin{mequation}
\label{eq:distortion_structure_appendix}
\|\bx\|^2 - L^2 \sum_{q=1}^{k}|\cC_q|\,\diamG(\cC_q)^2
\leq  \|\bPhif \ \bx\|^2 \leq \|\bx\|_2^2. 
\end{mequation}
\hfill$\square$
\end{proof}
\begin{corollary}
Let $L_q$ be the smoothness index inside cluster $C_q$, for all $q \in [k]$. 
This is the minimum $L_q$ such that:
\begin{equation*}
|\bx_i - \bx_j| \leq L_q \, \text{dist}_\cG(v_i,\, v_j), \quad \forall (i, j) 
\in C_q^2.
\end{equation*}
Then the following two inequalities hold:
\begin{mequation}
\begin{split}
\|\bx\|_2^2 - \sum_{q=1}^{k}|\cC_q| \sup_{\bx_i, \bx_j \in \bx_{\cC_q}} |\bx_i 
- \bx_j|_2^2 \leq & \\
\|\bx\|_2^2 - \sum_{q=1}^{k}L_q^2\,|\cC_q|\,\diamG(\cC_q)^2
\leq  & \|\bPhif \ \bx\|_2^2. 
\end{split}
\end{mequation}
\end{corollary}
\begin{proof}
As in Eq.\ref{eq:bound_inertia_cluster}, the second inequality is consequence 
of adding the local L-smoothness condition. 
\hfill$\square$
\end{proof}

\section{\label{sec:snr}Denoising properties}
In this section, we analyze the regularity condition of the signal of interest 
$\bs$, its relation with the noise and maximum cluster size.
Let $\bx$ be the acquired signal, which is a fixed signal of interest 
$\bs$  contaminated with an i.i.d. zero-mean Gaussian noise $\bn$ with 
variance $\sigma^2$, $\bx = \bs + \bn$.

With the purpose of ensuring clarity, we define $\bA = \bPhif\transpose \bPhif$.
Let $\MSE_{\text{appox}} = \expectation_{\bn}\left[\left\|\bs -  \bA \ 
\bx\right\|_2^2\right]$ and $\MSE_{\text{orig}} = 
\expectation_{\bn}\left[\|\bn\|_2^2\right]$ be the mean squared error with and 
without approximation, receptively.
As we are dealing with Gaussian noise, the risk of the raw data is 
$\MSE_{\text{orig}} = p \, \sigma^2$.

Given that $\|\bs\|_2^2$ is fixed, it is enough to show $\MSE_{\text{approx}} 
\leq \MSE_{\text{orig}}$ to ensure an increase in the SNR. 

\begin{proposition}
\label{prop:mse_approx}
Let $\bx = \bs + \bn \in \reals^{p}$ be an acquired signal, where $\bs$
is a fixed 
smooth L-Lipschitz signal and $\bn$ an i.i.d. zero-mean Gaussian 
noise with variance $\sigma^2$. 
Then, for a given grouping matrix $\bPhif\in\reals^{k\times p}$ the mean 
squared error of the approximation ($\MSE_{\text{approx}}$) is upper-bounded by
\begin{mequation}
\MSE_{\text{approx}} \leq  L^2 \sum_{q=1}^{k}|\cC_q|\,\diamG(\cC_q)^2 + 
\frac{k}{p}\, \MSE_{\text{orig}} .
\end{mequation}
\end{proposition}
\begin{proof}
We start by writing down the $\MSE_{\text{approx}}$, then we separate the 
components thanks to the i.i.d assumption and plug the upper-bound of the 
inertia, as follows
\begin{align*}
\MSE_{\text{approx}} & = \expectation_{\bn}\left[\left\|\bs -  \bA \ \bx 
\right\|_2^2\right]\\
& = \left\| (\bEye - \bA) \ \bs \right\|_2^2 + 
\expectation_{\bn}\left[\left\| \bA \  \bn \right\|_2^2\right]\\
& = \left\| (\bEye - \bA) \ \bs \right\|_2^2 + k\,\sigma^2\\
& \leq L^2 \sum_{q=1}^{k}|\cC_q|\,\diamG(\cC_q)^2 + k\,\sigma^2.
\end{align*}
\hfill$\square$
\end{proof}
\begin{corollary}
Let $\bx = \bs + \bn \in \reals^{p}$ be an acquired signal, where $\bs$ is a 
fixed 
a pairwise smooth L-Lipschitz signal and $\bn$ is an i.i.d. zero-mean Gaussian 
noise with variance $\sigma^2$. 
For a given grouping matrix $\bPhif\in\reals^{k\times p}$, the noise after 
approximation is reduced, $\MSE_{\text{approx}} \leq \MSE_{\text{orig}}$, only 
if the $L^2$ smoothness parameter satisfy 
\begin{mequation}
L^2 \leq \frac{(p - k)}{\sum_{q=1}^{k}|\cC_q|\,\diamG(\cC_q)^2}\, \sigma^2.
\end{mequation}
\end{corollary}
\begin{proof}
This is a direct result of the proposition\ref{prop:mse_approx} after some 
arithmetic manipulations, 
\begin{align*}
\MSE_{\text{approx}} & \leq L^2 \sum_{q=1}^{k}|\cC_q|\,\diamG(\cC_q)^2 + 
p\,\sigma^2 - (p - k)\,\sigma^2\\
& \leq L^2 \sum_{q=1}^{k}|\cC_q|\,\diamG(\cC_q)^2 + \MSE_{\text{orig}} - (p 
- k)\,\sigma^2.
\end{align*}
Then, to satisfy $\MSE_{\text{approx}} \leq \MSE_{\text{orig}}$, we must have: 
\begin{equation*}
L^2 \sum_{q=1}^{k}|\cC_q|\,\diamG(\cC_q)^2 \leq (p - k)\,\sigma^2,
\end{equation*}
which lead us to the upper-bound of the Lipschitz constant. 
\hfill$\square$
\end{proof}

\paragraph{\textbf{Cluster of the same size}}
This is a particular case, where we assume that the clusters $\cP = 
\{\cC_q\}_{q=1}^k$ have the same size, $\frac{p}{k}$. 
Under this assumption, the following holds:
\begin{mequation}
\label{eq:balanced_case}
\MSE_{\text{approx}} \leq p \left(\frac{L}{k}\right)^2  + 
\frac{k}{p}\, \MSE_{\text{orig}} = O\left(\max\left\{ 
\frac{p}{k^2},\frac{k}{p} \right\}\right)
.
\end{mequation}
The right-hand side of Eq.~\ref{eq:balanced_case} corresponds to 
the bias-variance tradeoff. Then, we need to balance these terms to 
maximize the rate of decay. 
Such balance can be achieved with
$k \sim p^{2/3}$ and hence $\MSE_{\text{approx}} \sim O(k^{-1/2})$. Such choice
for $k$ is consistent with values that we have found to work well
empirically.
\section{\label{sec:distortion} Experiment description: distortion}
To assess the quality of this approximation  (see 
Eq.2 and Eq.8, 
we randomly split half of the data to form a train and test clean signals 
$(\bS^{\text{train}}, \bS^{\text{test}})$ and a train corrupted data matrix 
$\bX^{\text{train}}$. 
We learn $\bPhi$ on the train corrupted data $\bX^{\text{train}}$. 
On the test data, we fit a proportionality constant $\eta$ that relates the 
distances in reductions of the corrupted data with the corresponding
distances in the clean signals.

We denote $\bdelta^{\text{orig}}_{(i,j)}$ the norm of the difference of the $i$ 
and $j$ uncorrupted signals, $\|\bS_{*, i}^{\text{test}} - 
\bS_{*, j}^{\text{test}}\|_2$, and  $\bdelta^{\text{noisy}}_{(i,j)}$ the
 norm of the difference of the $i$ and $j$ scaled noisy signals, $\eta 
 \|\bPhi\,\bX_{*, i}^{\text{test}} - \bPhi\,\bX_{*, j}^{\text{test}}\|_2$, 
 for all $(i, j) \in \left[\left\lfloor \frac{n}{2} \right\rfloor\right]^2$ 
 (note that $\bdelta \in \reals^{n (n - 1) / 
8}$).
We then use the relative distortion (RD) between $\bdelta^{\text{orig}}$ and 
$\bdelta^{\text{noisy}}$ to quantity the denoising effect of each method:
\begin{mequation}
\text{RD}(\bdelta^{\text{orig}}, \bdelta^{\text{noisy}})(\text{dB}) = 
-10\log_{10} \frac{\|\bdelta^{\text{noisy}} - 
\bdelta^{\text{orig}}\|_2^2}{\|\bdelta^{\text{orig}}\|_2^2}.
\end{mequation}
This measure gives us an insight on the distortion and possibly denoising 
effect. 
In particular, it shows us for which fraction of the signal the condition of 
Eq.7 is satisfied.

\section{Experiments}
\subsection{Empirical computational complexity}
We empirically assess the scalability of the algorithms as function of the 
input size. 
The test is carried out for a synthetic dataset composed of $10$ cubes. 
We varied their dimension $p \in \{8, 16, 64, 128\}^3$ and fix the number of 
clusters to $k = \left\lfloor \frac{p}{20} \right\rfloor$. 
We repeat $10$ times, and report the average computation time.\\

\begin{figure}[h!]
\centering
\includegraphics[width=0.91 \linewidth]{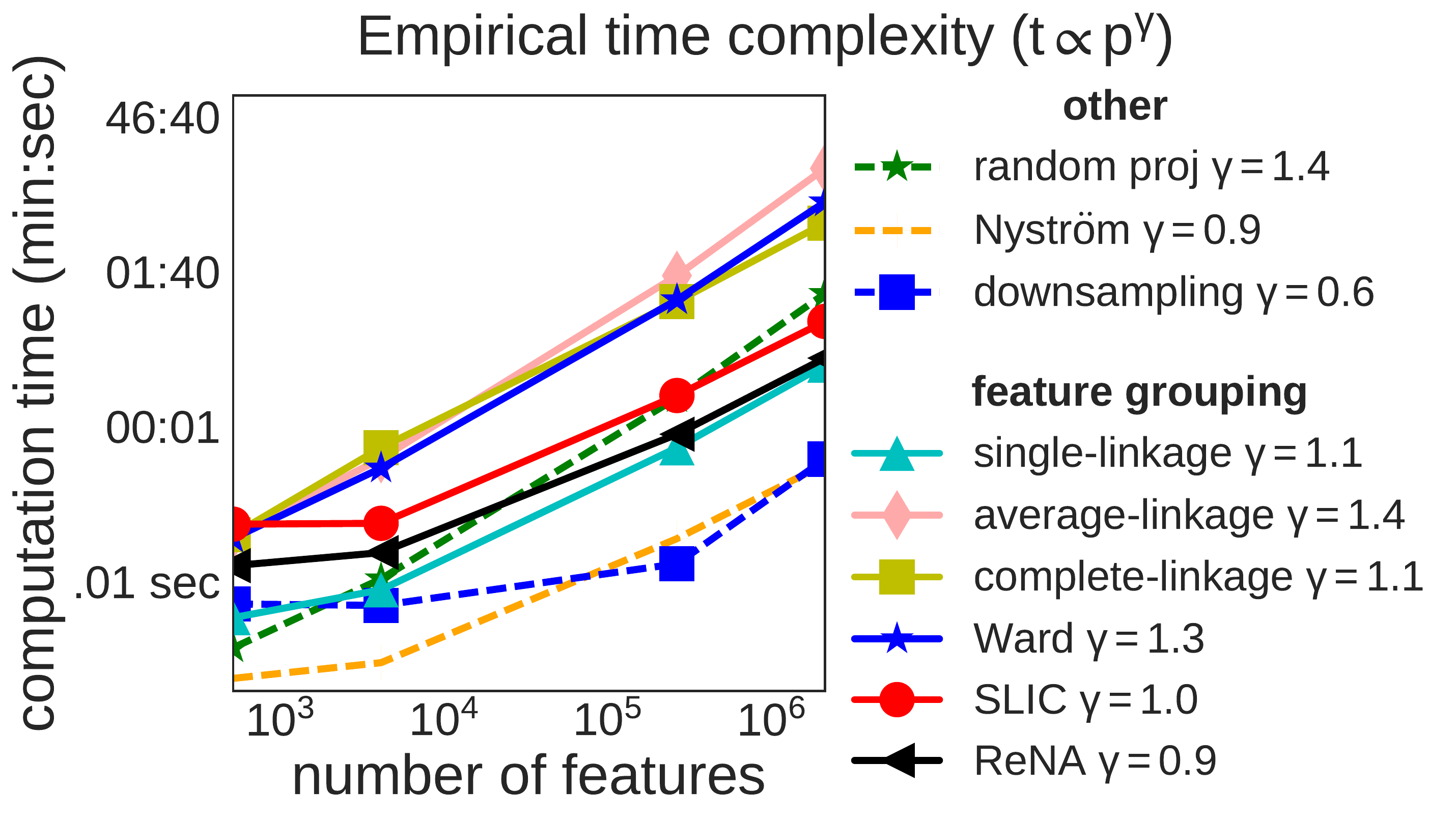} 
\caption{\textbf{Complexity of the computation time on a synthetic dataset:} 
Evaluation of the time complexity to find $\bPhi$ per algorithm on a synthetic 
data cube, for various feature space dimensions $p \in \{8, 16, 64, 128\}^3$ 
and fixing the number of clusters to $k = \left\lfloor \frac{p}{20} 
\right\rfloor $.
Downsampling displays a sub-linear time complexity, whereas 
Nystr\"om, single-linkage, complete-linkage, SLIC and ReNA present a linear 
behavior.
Complete-linkage, Ward and random projections have a sub-quadratic time 
behavior.
}
\label{fig:complexity}
\end{figure}

Fig.~\ref{fig:complexity} reports the computation time to estimate $\bPhi$ for 
the 
different approximation schemes. 
Nystr\"om, single-linkage, complete-linkage, SLIC and ReNA have a 
cost linear in the number $p$ of features.  
Random projections, average-linkage and Ward display a sub-quadratic time 
complexity, while downsampling has a sub-linear behavior.
Single-linkage and ReNA outperform other agglomerative methods, reducing
computation time by a factor of $10$. 
Profiling reveals that random projections' run time is dominated by
the random number generation. 
The scikit-learn implementation uses a Mersenne Twister algorithm, with
good entropic properties at the cost of increased
computations \cite{Katzgraber2010}. 

\subsection{Approximation of brain images with clustering-based dimension reduction}
\begin{figure*}[h!]
\centering \subfigure[Original]{\includegraphics[width=.1445
    \linewidth]{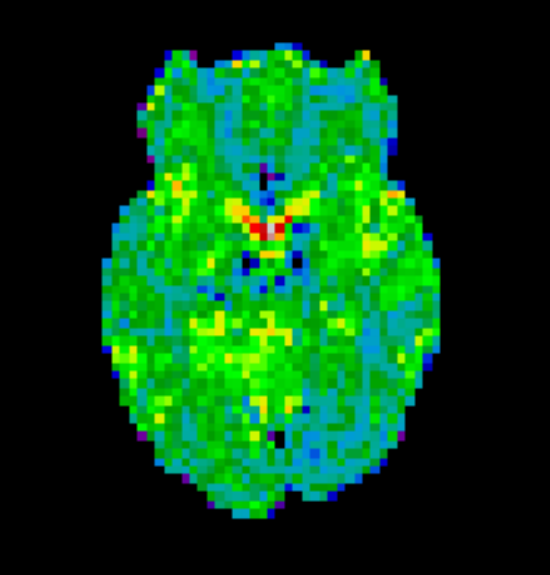}}\hspace{-3pt}
\subfigure[single-linkage]{\includegraphics[width=.1445
    \linewidth]{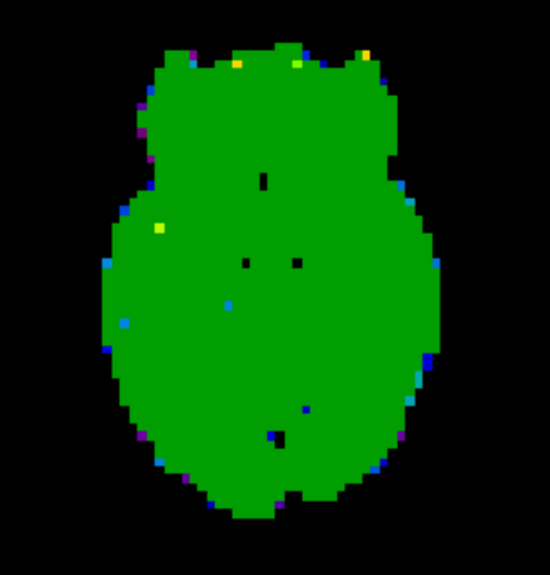}}\hspace{-3pt}
\subfigure[average-linkage]{\includegraphics[width=.1445
    \linewidth]{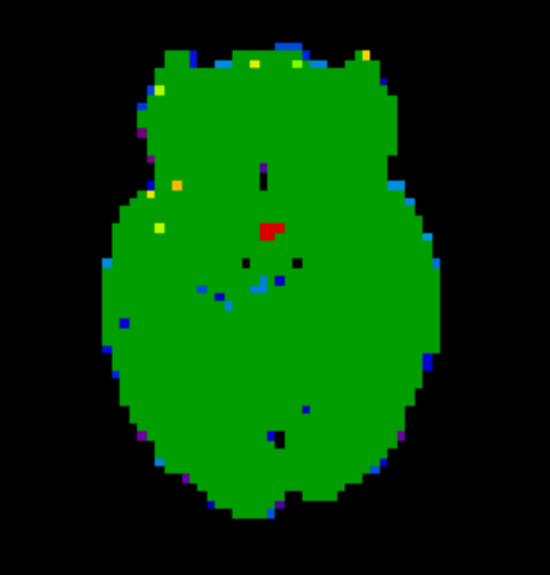}}\hspace{-3pt}
\subfigure[complete-linkage]{\includegraphics[width=.1445
    \linewidth]{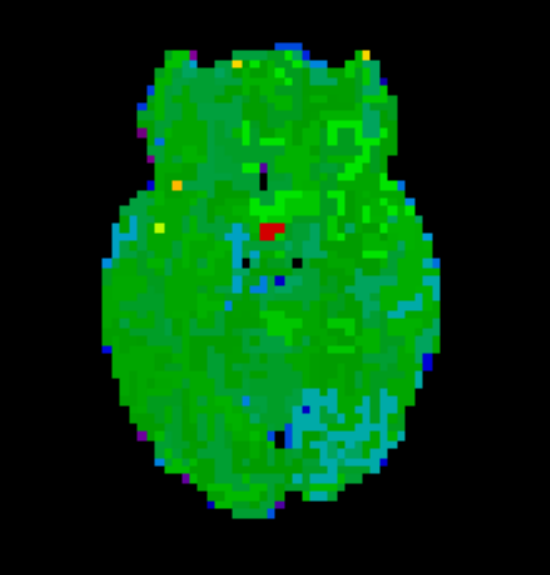}}\hspace{-3pt}
\subfigure[Ward]{\includegraphics[width=.1445
    \linewidth]{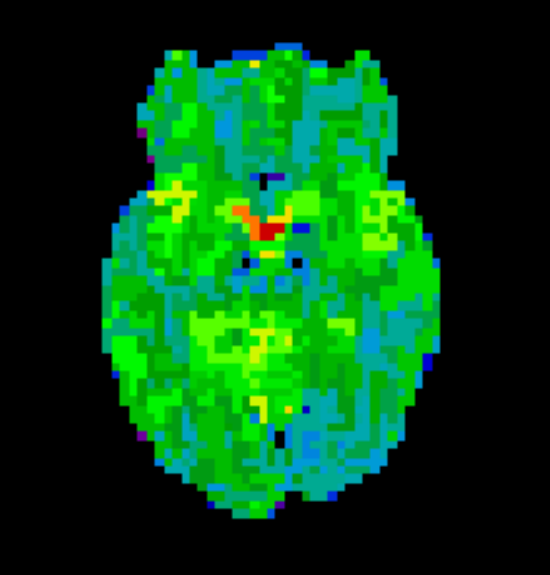}}\hspace{-3pt}
\subfigure[SLIC]{\includegraphics[width=.1445
    \linewidth]{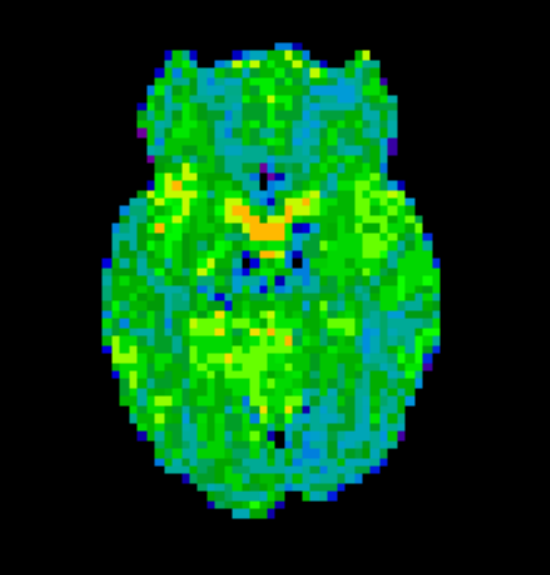}}\hspace{-3pt}
\subfigure[ReNA]{\includegraphics[width=.1445
    \linewidth]{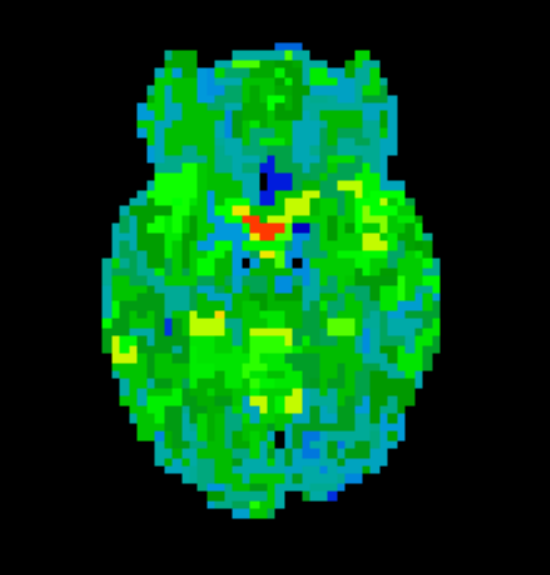}}\hspace{-3pt}
\caption{\textbf{Approximation of an MRI image obtained with various feature 
grouping algorithms:} A compressed representation of an MRI image (slice) using 
various clustering methods for a number of clusters $k=1000$.  Traditional 
agglomerative clustering methods exhibit giant clusters, losing meaningful 
information. 
By contrast, Ward, SLIC and ReNA algorithms present a better performance, 
finding balanced clusters.
}
\label{fig:example_compression}
\end{figure*}
Fig.~\ref{fig:example_compression} shows the approximations of
brain images using clustering methods.
It shows that traditional agglomerative clustering methods exhibit
giant clusters and lose meaningful information.
By contrast, Ward, SLIC and ReNA algorithms find balanced clusters,
and yield better approximations of the input.

\bibliographystyle{IEEEtran} 
\bibliography{biblio}